\documentclass[journal,twoside,web]{ieeecolor}
\usepackage{generic}
\usepackage{amsmath,amssymb,amsfonts}
\usepackage{graphicx}
\usepackage{textcomp}
\usepackage{booktabs}

\usepackage{bm}
\usepackage[ruled,linesnumbered]{algorithm2e}

\usepackage{amsthm}
\newtheorem{definition}{Definition}
\newtheorem{theorem}{Theorem}
\newtheorem{lemma}{Lemma}
\newtheorem{proposition}{Proposition}
\theoremstyle{remark}

\newtheorem{assumption}{Assumption}

\def\BibTeX{{\rm B\kern-.05em{\sc i\kern-.025em b}\kern-.08em
    T\kern-.1667em\lower.7ex\hbox{E}\kern-.125emX}}

\usepackage[colorlinks=false, citecolor=blue, urlcolor=blue]{hyperref}
\begin{document}
\title{A Forward-Inverse Dynamic Game Framework for Enhanced Multi-Agent
Trajectory Planning}
\author{Tianle Liu, Youcheng Niu, Jing Zeng,  Shuo Li, Jinming Xu$^*$
\thanks{The authors are with the College of Control Science and Engineering, Zhejiang University, Hangzhou 310027, China (e-mail: liutianle@zju.edu.cn; ycniu@zju.edu.cn; zengjing@zju.edu.cn; shuo.li@zju.edu.cn; jimmyxu@zju.edu.cn).}
\thanks{$^*$Corresponding author: Jinming Xu.}
}


\maketitle

\begin{abstract}
This paper studies feedback Nash equilibrium (FBNE) 
seeking for multi-agent trajectory planning in nonlinear 
dynamical systems with unknown agents' objectives and state-dependent inter-agent coupling. 
While dynamic game theory provides a principled framework 
for such problems, existing approaches typically assume fully rational agents with known objectives or rely on fixed regularization, limiting their ability to capture bounded rationality and spatially 
varying interaction intensity in safety-critical settings. To this end, we propose a KL-regularized 
dynamic game with a state-dependent weight that adaptively balances optimality and behavioral priors. To infer 
unknown cost parameters from demonstrated behaviors, we 
develop a context-aware inverse game module based on maximum-entropy inverse reinforcement learning with 
physics-informed regularization, ensuring structural consistency with the forward game. We establish per-iteration well-posedness of the 
regularized local game and show that the adaptive weighting function remains Lipschitz continuous under bounded nominal-trajectory updates. Numerical simulations 
and multi-robot experiments on cooperative navigation and 
merging scenarios validate the effectiveness of the proposed 
framework.
\end{abstract}


\section{Introduction}
\IEEEPARstart{P}lanning in multi-agent systems with strong 
interactions remains a fundamental challenge in safety-critical 
environments. Traditional “prediction-then-planning” methods neglect mutual interactions, often resulting in conservative or unsafe behaviors~\cite{gil2025predictability, liu2023learning}. 
Dynamic game theory addresses this limitation by explicitly 
modeling coupled interactions, and has shown promise in tasks 
involving safety constraints and long-horizon 
planning~\cite{li2023cost, 
lidard2024blending, rowold2024open, galati2025game}.

Existing game-theoretic trajectory planning 
methods~\cite{liu2023learning,peters2023online,li2023cost,lidard2024blending}, 
together with related studies on Nash equilibrium seeking in networked control 
systems~\cite{TCNS1,TCNS2,TCNS3,TCNS4}, where agents interact over coupled 
decision-making structures with partial information and uncertain nonlinear 
dynamics have substantially advanced equilibrium computation and closed-loop 
stabilization. Nevertheless, a key gap remains: unlike networked control 
systems where coupling is defined by a fixed communication graph, inter-agent 
coupling in physical environments arises from proximity and varies 
continuously over time, making it unclear whether the induced equilibrium 
remains behaviorally realistic when agents exhibit bounded rationality. Lidard 
et al.~\cite{lidard2024blending} take a step toward addressing this by 
introducing a Kullback--Leibler (KL) divergence term~\cite{ccinlar2011probability} to blend optimal control with data-driven 
priors. However, the trade-off coefficient is fixed per agent, 
leaving the balance between optimization and expert guidance invariant across 
interaction regimes. We therefore introduce a state-dependent mechanism that strengthens reference-policy adherence in high-risk regions and relaxes it elsewhere.

\begin{figure}[t]
      \centering
      \includegraphics[width=0.86\linewidth, height=0.48\linewidth]{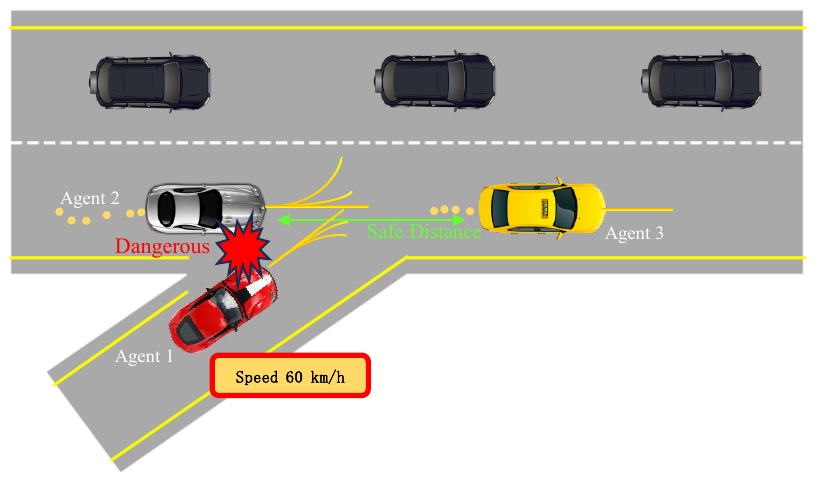}
      \caption{Red ego vehicle merging from an on-ramp onto a highway among four surrounding agents with uncertain intentions.}
      \label{figure1}
\end{figure}

A second major challenge is that many game-theoretic planning 
methods rely on prior knowledge of other agents' intentions, 
which is often unavailable due to sensor noise, incomplete 
observations, and unpredictable behaviors~\cite{le2021lucidgames}. 
As illustrated in Fig.~\ref{figure1}, a merging vehicle must 
reason about surrounding agents' latent goals while 
simultaneously planning its own trajectory. Inverse game 
theory has been employed to address this 
issue~\cite{mehr2023maximum,peters2023online}. Mehr et 
al.~\cite{mehr2023maximum} iteratively estimate cost weights 
by minimizing the discrepancy between the demonstrations and the 
learned policy. Donge et al.~\cite{TCNS_inverse} recover 
unknown cost functions via coupled optimal control and 
inverse-optimal control updates. Peters et 
al.~\cite{peters2023online} formulate goal inference as a 
non-convex optimization problem with KKT-based equilibrium 
constraints~\cite{KKT-condition}. Although promising, these 
methods focus primarily on recovering latent rewards or cost 
weights, while remaining limited in jointly modeling coupled 
multi-agent interactions and structured environmental context 
essential for interactive planning.

We propose a forward--inverse dynamic game framework for multi-agent trajectory planning under unknown agent objectives and state-dependent interaction risk. The forward game is formulated with a state-dependent KL regularization that adaptively 
balances optimality and reference-policy 
guidance. By freezing coefficients along the nominal 
trajectory, the resulting game admits a tractable linear--quadratic structure 
with guaranteed per-iteration well-posedness. The inverse game is formulated as a maximum-entropy inverse reinforcement
learning (IRL) problem. A scene-conditioned neural network jointly encodes
trajectory histories and map context to infer agent latent preferences by
estimating cost-function weights. In addition, a rule-aware regularization
term is incorporated to ensure consistency with the forward solver. The main contributions are summarized as follows.
\begin{itemize}
\item We propose a hybrid forward–inverse framework that integrates  game-theoretic modeling with learning from expert demonstrations for more realistic decisions. We further introduce an adaptive weighting scheme to balance data-driven influence, improving accuracy and contextual adaptability in trajectory planning.
\item We introduce a scene-aware neural network that is capable of jointly encoding multi-agent trajectories and maps to represent cost functions. To ensure robust and physically consistent reasoning, we propose a training mechanism that integrates maximum entropy IRL with physics-informed regularization terms.
\item We demonstrate the effectiveness of our algorithm over existing methods in intention inference and trajectory planning via simulated and real-world experiments in scenarios such as multi-agent collision avoidance navigation,  high-speed ramp merging, and position exchange.
\end{itemize}
\section{Related Work}
\subsection{Game-theoretic Motion Planning}
Game theory provides a framework for modeling non-cooperative trajectory planning, with applications in autonomous driving~\cite{fisac2019hierarchical,so2023mpogames}, robot navigation in crowds~\cite{galati2025game}, and shared control~\cite{11142955}. Game-theoretic trajectory planning methods aim to find a Nash equilibrium (NE) such that each agent's strategy is optimal given the other agents' strategies~\cite{bacsar1998dynamic}. Peters et al.~\cite{peters2023online} solve dynamic games using KKT conditions. Liu et al.~\cite{liu2023learning} model dynamic games as mixed complementarity problems (MCPs) to solve the open-loop NE. However, since their solutions are based on fixed trajectories and static variables, they cannot capture the state-dependent feedback strategies needed for a feedback Nash equilibrium (FBNE). FBNE~\cite{bacsar1998dynamic} enables agents to adjust their strategies dynamically according to real-time changes in system states, which aligns more closely with the interactive nature and robustness requirements of multi-agent systems~\cite{li2023cost}. However, since feedback strategies depend on the global structure of the state space, finding solutions can present challenges such as a high computational overhead~\cite{laine2023computation}. To solve these challenges, Li et al.~\cite{li2023cost} investigate FBNE solutions based on traditional optimal control theory; however, their approach does not account for the bounded rationality observed in human decision-making under real-world conditions. 

To address this issue, Mehr et al.~\cite{mehr2023maximum} develop a hybrid strategy that incorporates maximum entropy to attenuate the fully rational setting of an agent; however, they expand the nonlinear cost without including the cost term of control, resulting in a loss of accuracy. Lidard et al.~\cite{lidard2024blending} propose a dynamic game solution that combines data-driven priors with optimal control theory. However, their formulation adopts a
fixed $\lambda^i$ and does not provide theoretical support for the nonlinear
iterative extension beyond the linear-quadratic-Gaussian case. In this work,
we introduce a state-dependent regularization mechanism that adaptively
modulates this balance according to the local interaction context.
\subsection{Inverse Dynamic Games}
The inverse game problem aims to infer the underlying intentions of agents. This allows for more realistic interaction strategies that replicate expert behavior and provides valuable priors for future planning~\cite{mehr2023maximum, ruiz2025factorised}. Le Cleac'h et al.~\cite{le2021lucidgames} employ an unscented Kalman filter (UKF) to infer the cost functions of other agents and predict their behavior. However, their approach focuses on goal inference and does not solve the more challenging FBNE problem. Awasthi et al.~\cite{Awasthi_Lamperski_2020} reconstruct the agent cost functions by minimizing KKT residuals, which require full observations, limiting their practicality. 

Neural networks provide a powerful framework for modeling complex relationships between different terms of agents' cost functions. Finn et al.~\cite{Finn_Levine_Abbeel_2016} introduce a neural network-based cost function within a maximum entropy framework to model interactions among agents. Yu et al.~\cite{Yu_Song_Ermon_2019} develop an adversarial IRL approach that generalizes to high-dimensional environments with unknown dynamics, using deep neural networks and pseudolikelihood estimation. 
However, these methods have not been extended to multi-agent scenarios. More recently, Lidard et al.~\cite{lidard2024blending} use a fully connected network with data aggregation to learn cost-term weights. However, their method overlooks rich map information, limiting its ability to model agent interactions. In contrast, we leverage scene-aware contextual features to capture structural cues and model relationships among cost terms.

\section{Preliminaries and Problem Statement}\label{preliminaries}
In this section, we present the basic concepts and mathematical formulations of both the forward and inverse games. Specifically, we introduce the KL-divergence-regularized forward game formulation and the inverse game formulation for recovering unknown objectives from demonstrations.
\subsection{Forward Dynamic Games with KL Divergence}
We consider a game involving $N$ agents over a discrete time horizon $T$. We use a subscript to denote time and the superscript to indicate agents. For example, we use ${\mathbf{x}}_t^i \in {\mathbb{R}^{{m^i}}}$ and ${\mathbf{u}}_t^i \in {\mathbb{R}^{{n^i}}}$ to denote the state and control input of agent $i$ at time $t$, respectively. We let ${{\mathbf{x}}_t} = {[{\mathbf{x}}_t^{1, \top }, \ldots {\mathbf{x}}_t^{N, \top }]^ \top } \in {\mathbb{R}^m}$ denote the joint state with a dimension of $m = \sum {_i{m^i}}$, ${{\mathbf{u}}_t} = {[{\mathbf{u}}_t^{1, \top }, \ldots {\mathbf{u}}_t^{N, \top }]^ \top } \in {\mathbb{R}^n}$ denote the joint control with a dimension of $n = \sum {_i{n^i}} $. The joint state and control trajectories are denoted as $\mathbf{x}:=[\mathbf{x}_1,\ldots \mathbf{x}_T] ^ \top$, $\mathbf{u}:=[\mathbf{u}_1,\ldots \mathbf{u}_T] ^ \top$. Accordingly, the system dynamics is given by
\begin{equation}
{{\mathbf{x}}_{{t + 1}}} = f({{\mathbf{x}}_{t}},{{\mathbf{u}}_t}),
\label{dynamic}
\end{equation}
where $f:{\mathbb{R}^m} \times {\mathbb{R}^n} \to {\mathbb{R}^m}$ represents the dynamics of the underlying system.
Now, we are ready to introduce the concept of mixed strategies. At each time step, agent $i$ samples an action from a probability distribution $\pi_t^i(\mathbf{u}_t^i \mid \mathbf{x}_t)$. Each agent $i$ aims to minimize the cost function $J^i$, which evaluates the quality of its trajectory under applied control. Specifically, under the mixed strategy, $J^i$ is defined as follows:
\begin{equation}
{J^i}({\pi ^i})={\mathbb{E}_{\bm{\pi}}}[\sum\limits_{t = 1}^T {c_t^i({{\mathbf{x}}_t},{{\mathbf{u}}_t},{\boldsymbol{\omega}^i})\!+\!{\lambda ^i}{D_{KL}}(\pi _t^i(\left.  \cdot  \right|{{\mathbf{x}}_t})\left | {\tilde \pi _t^i (\left.  \cdot  \right|{{\mathbf{x}}_t})} \right.} ],
\label{cost}
\end{equation}
where $\pi ^i :=[\pi ^i_1, \ldots, \pi ^i_T ]$ denotes the sequence of strategies of agent $i$, $\bm{\pi} := (\pi^1, \ldots, \pi^N)$ denotes the profile of the joint 
strategy, $c_t^i$ is the original cost function of agent $i$ at time $t$, $\lambda^i$ is the scaling factor to balance the importance of the two cost items, $\boldsymbol{\omega}^i$ denotes the weights assigned to different components of the cost function and $D_{KL}$ denotes the KL divergence that captures the distance between the expert strategy $\tilde \pi _t^i$ and the current strategy $\pi _t^i$. Next, we introduce the concept of FBNE, as given in the following Definition~\ref{definition}.
\begin{definition}\label{definition}
(Feedback Nash Equilibrium~\cite{bacsar1998dynamic}) A policy $\pi^i$ is said to constitute a \textbf{FBNE} solution if no agent has an incentive to unilaterally alter the strategy, i.e.,
\[
J^i(\pi^{i*}, \pi^{\neg i*}) \leq J^i(\pi^i, \pi^{\neg i*}),
\quad \forall i \in [N],\ \forall \pi^i \in \Pi^i,
\label{nash}
\]
where $\neg i$ denotes all agents except $i$, and the superscript $(\cdot)^*$ denotes an indicator of optimality; $\Pi^i$ denotes the feedback strategy space of agent $i$.
\end{definition}
\subsection{Inverse Dynamic Games}\label{INV}
Solving the forward game requires that the cost functions $J^i$ of all 
agents are known a priori. In practice, however, agents' objectives are 
private, and must therefore be inferred from demonstrated 
behaviors. Let $\boldsymbol{\omega} := \{\boldsymbol{\omega}^i\}_{i \in [N]}$ 
denote the collection of cost weights for all agents. The inverse game 
problem is formulated as inferring $\boldsymbol{\omega}$ by maximizing the 
likelihood of an expert trajectory 
$\zeta = \{\bar{\mathbf{x}}_{[0:T]}, \bar{\mathbf{u}}_{[0:T]}\}$:
\begin{subequations}\label{eq:inv_obj}
\begin{align}
\max_{\boldsymbol{\omega}} \quad & P(\zeta \mid \boldsymbol{\omega}), 
\label{eq:inv_obj_5a} \\
\text{s.t.} \quad & \pi_{\boldsymbol{\omega}} \in \mathrm{FBNE}(\boldsymbol{\omega}),
\label{eq:inv_obj_5b}
\end{align}
\end{subequations}
where $P(\zeta \mid \boldsymbol{\omega})$ denotes the likelihood of observing 
the expert trajectory $\zeta$ under the equilibrium strategy induced by 
$\boldsymbol{\omega}$.

\begin{figure*}[htbp]
      \centering
      \includegraphics[width=0.87\linewidth, height=0.3\linewidth]{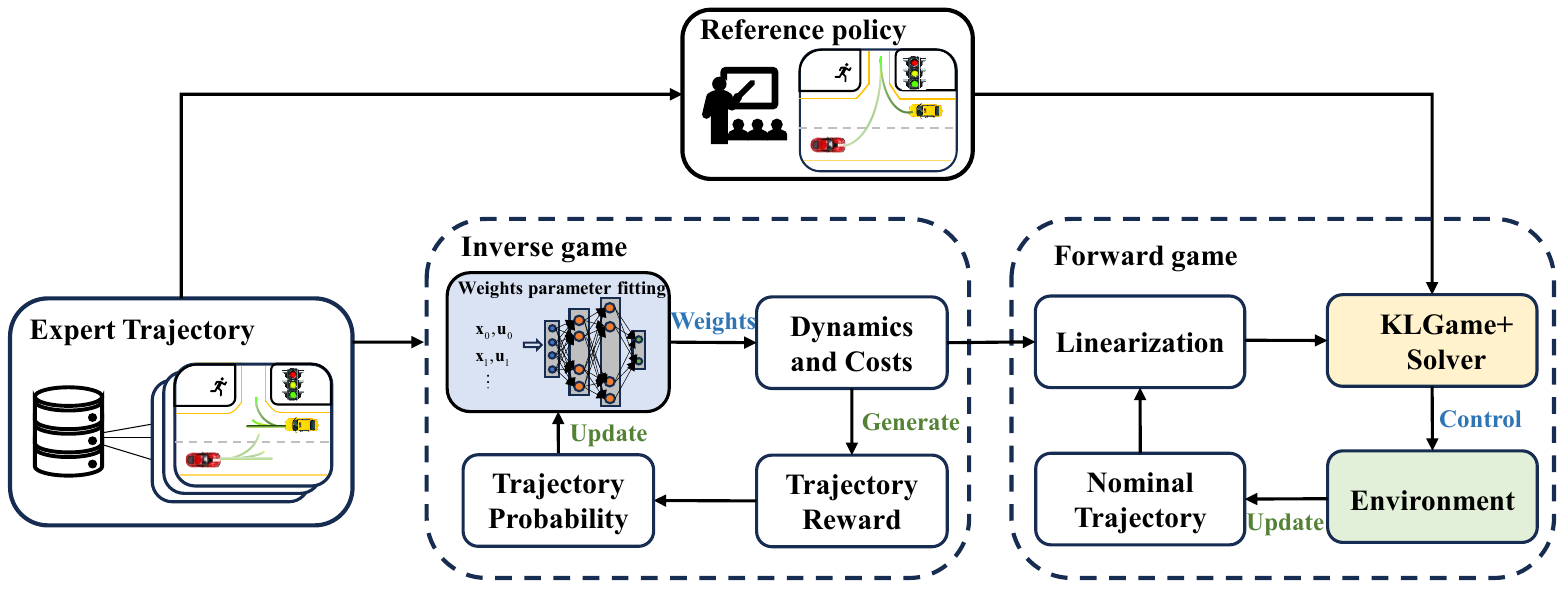}
      \caption{The proposed forward--inverse game framework. (1) In the inverse game, agent goals are inferred from expert data. (2) In the forward game, KL divergence is incorporated into the formulation, and the feedback Nash equilibrium under nonlinear dynamics is solved via iterative linearization.}
      \label{figure2}     
\end{figure*}
\section{The Proposed Framework}
We develop a framework that solves the FBNE problem using expert demonstrations. The forward game is solved with state-dependent weighting for nonlinear systems, while the inverse module infers agents’ intentions via deep maximum-entropy IRL with physics-informed constraints. Fig.~\ref{figure2} illustrates the general architecture of our approach.
\label{sec:Method}
\subsection{Forward Game Solver}
In the forward game, we linearize the nonlinear system around the nominal trajectory and then compute the forward FBNE. The nominal trajectory is then updated based on the obtained solution. The details are described in \textit{Algorithm~\ref{Ag1}}.
\begin{algorithm}
\SetAlgoLined
\KwIn{Initial state $\mathbf{x}_0$, expert trajectory $\zeta$, 
      time horizon $T$, convergence bound $\kappa_1$.}
\KwOut{Converged control $\hat{\mathbf{u}}^k_{0:T}$ and 
       trajectory $\hat{\mathbf{x}}^k_{0:T}$.}

$\eta^1 \leftarrow \zeta$,\ 
i.e.\ $\hat{\mathbf{x}}^1_{0:T}\leftarrow\bar{\mathbf{x}}_{0:T}$,\ 
$\hat{\mathbf{u}}^1_{0:T}\leftarrow\bar{\mathbf{u}}_{0:T}$\;

\For{iteration $k = 1, 2, \ldots$}{
  $\{\tilde{\pi}^i(\cdot|\mathbf{x})\} \leftarrow 
   LaplaceApproximation(\eta^k)$\;

  $\{A_t, B_t^i\} \leftarrow Linearize(\eta^k)$\;

  $\{l_t^i, Q_t^i, R_t^{ij}, r_t^{ij}\} \leftarrow 
   Quadraticize(\eta^k)$\;

  $\lambda_t^{i,k} \leftarrow GetScaleParameter(\eta^k)$\;

  $\mu_t^{i,*}(\mathbf{x}_t;\eta_t^k) \leftarrow 
   KLGame{+}Solver(A_t, B_t^i, l_t^i, Q_t^i, 
                   R_t^{ij}, r_t^{ij}, \lambda_t^{i,k})$\;

  $\hat{\mathbf{u}}^{k+1} \leftarrow 
   stepToward(\hat{\mathbf{u}}^k,\, 
   \mu_t^{i,*}(\mathbf{x}_t;\eta_t^k))$\;

  \lIf{$\|\eta^k - \eta^{k-1}\| < \kappa_1$}
      {return $\hat{\mathbf{u}}^k$, $\hat{\mathbf{x}}^k_{0:T}$}
}
\caption{Forward Game Solver}
\label{Ag1}
\end{algorithm}

\textbf{Linearization}: In practical trajectory planning, both the dynamics and cost functions are often nonlinear. Following~\cite{fridovich2020efficient}, we approximate them by local quadratic expansions around a nominal trajectory, as in \textit{Lines 4--5 of Algorithm~\ref{Ag1}}. Specifically, we linearize the system dynamics and cost around
$\eta^k := \{\hat{\mathbf{x}}^k_{0:T},\, \hat{\mathbf{u}}^k_{0:T}\}$
at the $k$-th outer iteration, with $\eta^1 \leftarrow \zeta$, yielding:
    \begin{equation}
        \delta {{\mathbf{x}}_{t + 1}} \approx {A_t}\delta {{\mathbf{x}}_t} + \sum\limits_{j \in [N]} {B_t^j} \delta {\mathbf{u}}_t^j,
    \label{eq:4}
    \end{equation}
where $\delta \mathbf{{x}_{t}} = \mathbf{{x}}_t - {\mathbf{\hat x}_t}$, $\delta \mathbf{u}_{t}^j = \mathbf{u}_t^j - \mathbf{\hat u}_t^j$, and ${A_t} = {D_\mathbf{x}}f( \cdot )$, $B_t^j = {D_{\mathbf{u}_t^j}}f( \cdot )$ are the Jacobians of the system dynamics in Eq.~\eqref{dynamic}. We further obtain a quadratic approximation of the cost function for each agent. For each agent $i \in [N]$ we get
\begin{gather}
c_t^i\left( {{{\mathbf{x}}_t},{{\mathbf{u}}_t},{\boldsymbol{\omega }^i}} \right) \approx c_t^i\left( {{{{\mathbf{\hat x}}}_t},{{{\mathbf{\hat u}}}_t},{\boldsymbol{\omega }^i}} \right) + \frac{1}{2}(\delta {\mathbf{x}}_t^\top Q_t^i + 2l_t^{i,\top})\delta {{\mathbf{x}}_t} \notag \\+ \frac{1}{2}\sum\limits_{j \in [N]} {(\delta {\mathbf{u}_t^{j,\top}} R_t^{ij} + 2r_t^{ij,\top})} {\text{ }}\delta {{\mathbf{u}_t^j}},
\label{eq:5}
\end{gather}
where $l_t^i \triangleq D_{\hat{\mathbf{x}}^k} c_t^i$ and 
$r_t^{ij} \triangleq D_{\hat{\mathbf{u}}^{j,k}} c_t^i$ are the 
first-order derivatives of agent $i$'s stage cost with respect 
to the state and agent $j$'s control, respectively, and 
$Q_t^i \triangleq D^2_{\hat{\mathbf{x}}^k\hat{\mathbf{x}}^k} c_t^i$ 
and $R_t^{ij} \triangleq D^2_{\hat{\mathbf{u}}^{j,k}\hat{\mathbf{u}}^{j,k}} c_t^i$ 
are the corresponding Hessian matrices. In the frozen local 
linear--quadratic game around the nominal trajectory, we assume 
$Q_t^i \succeq 0$, $R_t^{ij} \succeq 0$, and $R_t^{ii} \succ 0$. 
Mixed partials $D^2_{\hat{\mathbf{x}}^k \hat{\mathbf{u}}^{j,k}}$ 
and $D^2_{\hat{\mathbf{u}}^{i,k}\hat{\mathbf{u}}^{j,k}}$ have been omitted for simplicity, but can be incorporated if needed.

We construct the reference policy using a game-theoretic Boltzmann distribution~\cite{jeongfinite}, where the soft Q-function is approximated by the cumulative cost along the nominal trajectory: $\mathcal{Q}^i(\mathbf{x}_t, \mathbf{u}_t) \approx \sum_{s=t}^{T} c_s^i(\mathbf{x}_s, \mathbf{u}_s, \boldsymbol{\omega}^i)$. The resulting expert policy is given by Eq.~\eqref{eq:6a} and used in \textit{Line 3 of Algorithm~\ref{Ag1}}, with the temperature parameter $\tau$ controlling the policy entropy and normalization $Z^i(\mathbf{x})$ defined in Eq.~\eqref{6b}.
\begin{subequations}
\begin{equation}
\tilde{\pi}^i(\mathbf{u} \mid \mathbf{x}) = \frac{1}{Z^i(\mathbf{x})} \exp\left( -\frac{1}{\tau} \mathcal{Q}^i(\mathbf{x}, \mathbf{u}, \boldsymbol{\omega}^i) \right),
\label{eq:6a}
\end{equation}
\begin{equation}
Z^i(\mathbf{x}) = \int \exp\left( -\frac{1}{\tau} \mathcal{Q}^i(\mathbf{x}, \mathbf{u}, \boldsymbol{\omega}^i) \right) d\mathbf{u}.
\label{6b}
\end{equation}
\end{subequations}

To derive a tractable reference policy, at each outer iteration 
$k$ we apply a Laplace approximation~\cite{bishop2006pattern} 
to $\tilde{\pi}^i$. Specifically, we perform a second-order Taylor expansion of \(\mathcal{Q}^i\), equivalently of \(\ln\tilde{\pi}_t^i\) up to scaling, around the current nominal 
action $\hat{\mathbf{u}}_t^{i,k}$ of agent $i$, assuming the 
nominal state $\hat{\mathbf{x}}_t^k$ is fixed.
This yields a local Gaussian approximation:
\begin{equation}
    \tilde{\pi}_t^i(\mathbf{u}_t^i \mid \mathbf{x}_t) \approx 
    \mathcal{N}\!\left(
    \tilde{\mu}_t^i(\eta_t^k),\; 
    \tilde{\Sigma}_t^i(\eta_t^k)
    \right),
\label{eq:expert_policy}
\end{equation}
where $\eta_t^k := \{\hat{\mathbf{x}}_t^k,\, \hat{\mathbf{u}}_t^k\}$. The mean and covariance are given by
\begin{subequations}
\begin{equation}
    \tilde{\mu}_t^i(\eta_t^k) := \hat{\mathbf{u}}_t^{i,k},
\label{eq:mean}
\end{equation}
\begin{equation}
    \tilde{\Sigma}_t^i(\eta_t^k) := 
    -\left[\nabla_{\mathbf{u}_t^i}^2 \ln \tilde{\pi}_t^i
    (\mathbf{u}_t^i \mid \mathbf{x}_t)
    \Big|_{\mathbf{u}_t^i = \hat{\mathbf{u}}_t^{i,k}}
    \right]^{-1}.
\label{eq:cov}
\end{equation}
\end{subequations}

We assume that the local curvature satisfies
\(-\nabla_{\mathbf{u}_t^i}^2 \ln \tilde{\pi}_t^i
(\mathbf{u}_t^i \mid \mathbf{x}_t)|_{\mathbf{u}_t^i=\hat{\mathbf{u}}_t^{i,k}}\succ0\),
which ensures \(\tilde{\Sigma}_t^i(\eta_t^k)\succ0\).
Setting the mean directly to the nominal control as in Eq.~\eqref{eq:mean} avoids
the functional optimization subproblem required by the original KLGame
implementation~\cite{lidard2024blending}.

\textbf{KLGame+ Solver}:
We augment the cost function as in Eq.~\eqref{cost} with a KL divergence term and introduce a state-aware mechanism to balance task performance and data-driven behavior. The following lemma provides the key structural property underlying the analytic characterization of the KL-regularized optimal policy and Theorem~\ref{th1}.

\begin{lemma}[Regularized Variational 
Formula~\cite{lidard2024blending}]
\label{lemma:variational}
Let \(\mathcal{Q}:\mathcal{X}\times\mathcal{U}\to\mathbb{R}\) be a bounded measurable function. 
Then, for any \(\lambda>0\), reference probability measure
\(\tilde{\pi}(\cdot|\mathbf{x})\), and state \(\mathbf{x}\in\mathcal{X}\), we have
\begin{equation}
\label{eq:variational}
\begin{split}
     -\lambda \log \!&\int_{\mathcal{U}} 
     e^{-\mathcal{Q}(\mathbf{x},\mathbf{u})/\lambda} 
     \tilde{\pi}(\mathbf{u}|\mathbf{x})\,\mathrm{d}\mathbf{u} \\
    &=
    \inf_{\pi}
    \left\{
    \mathbb{E}_{\pi}[\mathcal{Q}(\mathbf{x},\mathbf{u})]
    + \lambda D_{\mathrm{KL}}(\pi \| \tilde{\pi})
    \right\}.
\end{split}
\end{equation}
The infimum is uniquely achieved by the Boltzmann policy
\begin{equation}
    \pi^{*}(\mathbf{u}|\mathbf{x})
\propto
\tilde{\pi}(\mathbf{u}|\mathbf{x})
\exp\!\bigl(-\mathcal{Q}(\mathbf{x},\mathbf{u})/\lambda\bigr).
\end{equation}
Moreover, if \(\tilde{\pi}\) is Gaussian and \(\mathcal{Q}\) is quadratic in \(\mathbf{u}\), then \(\pi^*\) is also Gaussian with computable mean and covariance.
\end{lemma}

\begin{proof}
See Appendix \ref{lemma1_proof}.    
\end{proof}
Next, Theorem~\ref{th1} derives the Gaussian feedback strategy used by the
KLGame+ solver in Line~7 of Algorithm~\ref{Ag1}.

\begin{theorem}\label{th1}
   The $N$-agent dynamic game with KL divergence admitsadmits a feedback Nash equilibrium solution if the system dynamics and cost function can be linearized into the forms of Eqs.~\eqref{eq:4}--\eqref{eq:5}, and the reference policies are Gaussian as in Eq.~\eqref{eq:expert_policy}. The solution is a set of time-varying policies $\pi _{t}^{i,*}\sim \mathcal{N} (\mu _{t}^{i*},\Sigma _{t}^{i*})$, with mean and covariance given by
    \begin{subequations}
        \begin{equation}
            \mu _t^{i*} =  - P_t^i{\delta \mathbf{x}_t} - \alpha _t^i,
        \end{equation}
        \begin{equation}
        \label{conv1}
            \Sigma _t^{i*} = {\left[ {\frac{1}{{{\lambda ^i}}}\left( {R_t^{ii} + B_t^{{i^\top}}Z_{t + 1}^iB_t^i} \right) + {{\left( {\tilde \Sigma _t^i} \right)}^{ - 1}}} \right]^{ - 1}},
        \end{equation}
    \end{subequations}
where the matrices $P_t^i$ and the vectors $\alpha_t^i$ satisfy the coupled linear systems in Eqs.~\eqref{13a}--\eqref{13b}:
\begin{subequations}
\setlength{\jot}{1pt}
    \begin{align}
       \left[ {R_t^{ii} + {\lambda ^i}{{\left( {\tilde \Sigma _t^i} \right)}^{ - 1}} + B_t^{i \top }Z_{t + 1}^iB_t^i} \right]P_t^i + B_t^{i \top }Z_{t + 1}^i \cdot\notag \\\sum\limits_{j \in [N],\, j \neq i} {B_t^j} P_t^j = B_t^{i \top }Z_{t + 1}^i{A_t},
    \label{13a}
    \end{align}
    \begin{align}
    \left[ {R_t^{ii} + {\lambda ^i}{{\left( {\tilde \Sigma _t^i} \right)}^{ - 1}} + B_t^{i \top }{Z_{t + 1}^i}B_t^i} \right]\alpha _t^i + B_t^{i \top }Z_{t + 1}^i\cdot \notag \\\sum\limits_{j \in [N],\, j \neq i} {B_t^j} \alpha _t^j = B_t^{i \top }z_{t + 1}^i + r_t^{ii} - {\lambda ^i}{\left( {\tilde \Sigma _t^i} \right)^{ - 1}}\tilde \mu _t^i,
    \label{13b}
    \end{align}
\end{subequations}
and the parameters of the value function $(Z_t^i,z_t^i)$ are computed recursively backward in time, as shown in Eqs.~\eqref{14a}-\eqref{14b}:
\begin{subequations}\allowdisplaybreaks
\setlength{\jot}{2pt}
\begin{align}
Z_t^i &= Q_t^i +\sum\limits_{j \in [N]} {{{\left( {P_t^j} \right)}^\top}} R_t^{ij}P_t^j + F_t^\top Z_{t + 1}^i{F_t} + \notag \\
&\quad {\lambda ^i}P_t^{i \top }{\left( {\tilde \Sigma _t^i} \right)^{ - 1}}P_t^i,\quad Z_T^i = Q_T^i,
\label{14a}\\
z_t^i &= l_t^i+\sum\limits_{j \in [N]} {{{\left( {P_t^j} \right)}^{\top}}} R_t^{ij}\alpha _t^j + F_t^{\top}\left( {z_{t + 1}^i + Z_{t + 1}^i{\beta _t}} \right) \notag \\
&\quad  \!-\!\sum_{j \in [N]} P_t^{j\top} r_t^{ij}+ {\lambda ^i}P_t^{i,{\top}}{\left( {\tilde \Sigma _t^i} \right)^{ - 1}}\left( {\alpha _t^i \!-\! \tilde \mu _t^i} \right), \quad z_T^i = l_T^i,
\label{14b}
\end{align}
\end{subequations}
where ${F_t} = {A_t} - \sum {_{j \in [N]}B_t^jP_t^j} $ and ${\beta _t} =  - \sum {_{j \in [N]}B_t^j\alpha _t^j}$.
\end{theorem}

\begin{proof}
See Appendix \ref{sec-proof-th1}.    
\end{proof}
In Section~\ref{preliminaries}, \(\lambda^i\) is treated as a constant for simplicity. The original KLGame framework~\cite{lidard2024blending} uses fixed \(\lambda^i\), which cannot capture adherence in safety-critical regions and weaker adherence in lower-risk ones. We therefore generalize \(\lambda^i\) to a state-dependent function \(\lambda^i(\mathbf{x}_t)\) under the following assumption.
\begin{assumption}\label{ass:regularity}
For every agent $i\in[N]$:
\textnormal{(A1)}~the static obstacle~$\mathcal{O}\subset\mathbb{R}^{2}$ is non-empty and closed;
\textnormal{(A2)}~the nominal trajectory is collision-free, i.e.,
$\hat{p}_t^{i,k} \notin \mathcal{O}$ and
$\hat{p}_t^{i,k} \neq \hat{p}_t^{j,k}$ for all $k,t,j\neq i$;
\textnormal{(A3)}~the position extraction map
$\mathcal{P}^{i}:\mathbb{R}^{m}\to\mathbb{R}^{2}$, defined by
$p^{i}=\mathcal{P}^{i}\mathbf{x}$, selects the position coordinates of
agent $i$ from the joint state vector and is linear and non-expansive,
i.e., $\|\mathcal{P}^{i}\|_{\mathrm{op}}\le 1$, where
$\|\cdot\|_{\mathrm{op}}$ denotes the operator norm. \textnormal{(A4)}~For each outer iteration $k$ and time step $t$, the
coupled linear systems for $\{P_t^i\}_{i\in[N]}$ and
$\{\alpha_t^i\}_{i\in[N]}$ in Eqs.~\eqref{13a}--\eqref{13b}, with the
frozen coefficient $\lambda_t^{i,k}$, are nonsingular.
\end{assumption}

For each agent $i \in [N]$, we define
\begin{equation}
\label{eq:lambda_state}
\lambda^i(\mathbf{x}_t) = \lambda_{\min}^i + (\lambda_{\max}^i - \lambda_{\min}^i) \rho^i(\mathbf{x}_t),
\end{equation}
where
\begin{equation}
\label{eq:rho_state}
\rho^i(\mathbf{x}_t) = \exp\!\left(-\frac{(d_{\mathrm{obs}}^i(\mathbf{x}_t))^2}{2\sigma^2}\right),
\end{equation}
with
$d_{\mathrm{obs}}^{i}(\mathbf{x}_t)
:= \min\!\bigl(
     \mathrm{dist}(p_t^{i},\mathcal{O}),\;
     \min_{j\neq i}\|p_t^{i}-p_t^{j}\|
   \bigr)$.
Let $\eta^k:=\{\hat{\mathbf{x}}_{0:T}^k,\hat{\mathbf{u}}_{0:T}^k\}$ denote the nominal trajectory at the $k$-th outer iteration. The state-dependent coefficient is evaluated along $\eta^k$ by setting
\begin{equation}
\label{eq:frozen_lambda}
\lambda_t^{i,k}:=\lambda^i(\hat{\mathbf{x}}_t^k), \qquad t\in[T],\ i\in[N],
\end{equation}
and the sequence \(\{\lambda_t^{i,k}\}\) is fixed during each inner Riccati recursion. Hence, each outer iteration solves a \emph{frozen-regularization local game}, with the state dependence of \(\lambda^i\) appearing only across outer iterations. The following two propositions establish the well-posedness of this frozen local game and the Lipschitz continuity of the state-dependent regularization coefficient, respectively.
 
\begin{proposition}\label{prop:lambda_wellposed}
Consider the system in Eq.~\eqref{dynamic} under the cost 
function in Eq.~\eqref{cost}. Suppose Assumption~\ref{ass:regularity} holds. 
Let the local dynamics and cost be locally approximated 
as in Eqs.~\eqref{eq:4}--\eqref{eq:5} with $Q_t^i\succeq 0$, 
$R_t^{ij}\succeq 0$, $R_t^{ii}\succeq r_{\min}^{i}I\succ 0$, 
and $\tilde{\Sigma}_t^{i}\succ 0$ for all $t\in[T]$ and 
$i,j\in[N]$. Further, let the regularization coefficients be fixed at 
each outer iteration $k$ as in Eq.~\eqref{eq:frozen_lambda}, and suppose
that the coupled linear system in Eqs.~\eqref{13a}--\eqref{13b} is 
nonsingular at each time step. Then, for all $t\in[T]$, $i\in[N]$, and 
$k\geq0$, the following statements hold:
\begin{enumerate}
     \item $Z_t^i\succeq 0$.
    \item $M_t^{i,k}\succeq m_0^{i}I\succ 0$ uniformly
    in $k$ and $t$, where
    \begin{equation}\label{eq:M_lambda}
    M_t^{i,k}
    := R_t^{ii}
     + \lambda_t^{i,k}(\tilde{\Sigma}_t^{i})^{-1}
     + B_t^{i\top}Z_{t+1}^{i}B_t^{i}.
    \end{equation}
    \item The covariance update is well-defined and satisfies
    $\Sigma_t^{i*}\succ 0$. Moreover, the corresponding Gaussian FBNE is unique.
\end{enumerate}
\end{proposition}
\begin{proof}
See Appendix \ref{proof_pro2}.    
\end{proof}

\begin{proposition}\label{prop:lambda_regular}
Suppose Assumption~\ref{ass:regularity}~(A1) and~(A3) hold,
$\sigma>0$, and $0<\lambda_{\min}^i\le\lambda_{\max}^i$. Then the
state-dependent coefficient $\lambda^i$ defined in
Eq.~\eqref{eq:lambda_state} maps $\mathbb{R}^m$ into
$[\lambda_{\min}^i,\lambda_{\max}^i]$ and is Lipschitz continuous, i.e.,
there exists a constant $C_\lambda^i>0$ such that
\begin{equation}
|\lambda^i(\mathbf{x})-\lambda^i(\mathbf{x}')|
\le
C_\lambda^i\|\mathbf{x}-\mathbf{x}'\|,
\qquad
\forall \mathbf{x},\mathbf{x}'\in\mathbb{R}^m .
\end{equation}
\end{proposition}

\begin{proof}
See Appendix \ref{proof_pro1}.    
\end{proof}
Propositions \ref{prop:lambda_wellposed} and \ref{prop:lambda_regular} establish that the
state-dependent regularization preserves per-iteration
solvability with a uniform bound~$m_0^{i}$ independent
of~$k$, and evolves with the Lipschitz
constant~$C_\lambda^{i}$.
Moreover, since $\lambda_t^{i,k}=\lambda^i(\hat{\mathbf{x}}_t^k)$ and
$\lambda^i(\cdot)$ is Lipschitz continuous, the frozen coefficients
$\lambda_t^{i,k}$ also converge whenever the nominal trajectory
sequence $\hat{\mathbf{x}}_t^k$ converges. The
Algorithm~\ref{Ag1} therefore remains within the sequential
local LQ framework of~\cite{lidard2024blending}.

\textbf{Nominal Trajectory Update}: Regarding the method for updating the nominal trajectory $\eta^k$, we adopt the form of the control strategy mean, i.e. $\delta {\mathbf{u}}_t^i =\varepsilon\ \mu _t^{i,*}({{\mathbf{x}}_t};{{ \eta }_t}) =  -\varepsilon P_t^i\delta {{\mathbf{x}}_t} - \varepsilon \alpha _t^i$, $\varepsilon  \in (0,1]$, which is used to employ a linear search strategy to prevent the updated trajectory from deviating excessively from the nominal trajectory $\eta^k$ to avoid divergence of the algorithm. The updated strategy $\mathbf{u}_t^i = \hat{\mathbf{u}}_t^{i,k} + \delta\mathbf{u}_t^{i,*}$ is then used to update the nominal trajectory from the same starting state in \textit{Line 8 of Algorithm~\ref{Ag1}}. This update procedure is repeated until convergence as described in \textit{Line 9 of Algorithm~\ref{Ag1}}, i.e., when the deviation between the newly generated state trajectory and the previous one falls below a predefined threshold.
\subsection{Inverse Game Solver} \label{inverse game solver}
 In this section, we present the inverse game solver combined with a neural network that infers the intentions of other agents to solve the problem posed in Section~\ref{INV}. Details of the algorithm used can be found in \textit{Algorithm~\ref{Alg2}}.
\begin{algorithm}
\caption{Inverse Game Solver}
\label{Alg2}
\KwIn{Dynamics $f$, Map information $c$,  Learning rate $\alpha$,  Expert trajectory set $\mathcal{D} = \{\zeta_k\}_{k=1}^K$, Cost primitives $\mathbf{f}^i$ for agent $i$, Rule loss weight $\nu_{\text{rule}}$, Convergence bound $\kappa_2$,  Sampling horizon $T$}
\KwOut{Cost weights $\omega^{i*} = \phi_\theta^i(\zeta_k, c)$ for agent $i$}
Initialize network parameters $\theta^0$, a buffer $\mathcal{B} \gets \left[ \,\, \right]$.\\
Compute expert feature expectation: $\mathcal{B} \gets {\mathbf{f}}_{\zeta_k}, \forall k \in [K]$.\\
\ForEach{expert trajectory $\zeta_k \in \mathcal{D}$}{
Generate a trajectory set $\tilde{\mathcal{D}}_k = \left\{ \tilde{\zeta}_{k}^{j} \right\}$ with the same initial state as $\zeta_k$ according to the sampling horizon $T$.\\
\ForEach{$\tilde{\zeta}_{k}^{j} \in \tilde{\mathcal{D}}_k$}{
Calculate the feature vector: $\mathbf{f}^i_{\tilde{\zeta}_{k}^{j}}$.\\
Add them to buffer: $\mathcal{B} \gets \tilde{\zeta}_{k}^{j}, \mathbf{f}_{\tilde{\zeta}_{k}^{j}}$.\\
}
}
\While{not converged {$\|\theta^{(\ell+1)} - \theta^\ell\| \geq \kappa_2$}}{
Obtain agent $i$'s weights: $\boldsymbol{\omega}^i = \phi_\theta^i(\zeta_k, c), \quad \forall \zeta_k \in \mathcal{D}$.\\
Calculate the trajectory reward feature from $\mathcal{B}$:\\
\quad $R_{\boldsymbol{\omega}^i}(\zeta_k) = {\boldsymbol{\omega}^i}^\top \mathbf{f}^i_{\zeta_k}$, \quad $R_{\boldsymbol{\omega}^i}(\tilde{\zeta}_k^j) = {\boldsymbol{\omega}^i}^\top \mathbf{f}^i_{\tilde{\zeta}_k^j}$.\\
Compute MaxEnt IRL loss: $\mathcal{L}_{\mathrm{MaxEnt}}^{i}$.\\
Compute rule-based regularization term: $\mathcal{L}_{\mathrm{rule}}^{i}$.\\
Combine total loss: \\
\quad $\mathcal{L}_{\text{total}}^i = \mathcal{L}_{\text{MaxEnt}}^i + \nu_{\text{rule}} \cdot \mathcal{L}_{\text{rule}}^i$.\\
Update network parameters: \\
\quad $\theta^{(\ell+1)} \leftarrow \theta^{\ell} - \alpha \cdot \nabla_\theta \mathcal{L}_{\text{total}}^i$.\\
}
\Return{weights $\omega_i^* = \phi_{\theta^i}(\zeta_k, c)$ for agent $i$}
\end{algorithm}

\begin{figure}[htbp]
      \centering
\includegraphics[width=0.88\linewidth,height=0.36\linewidth]{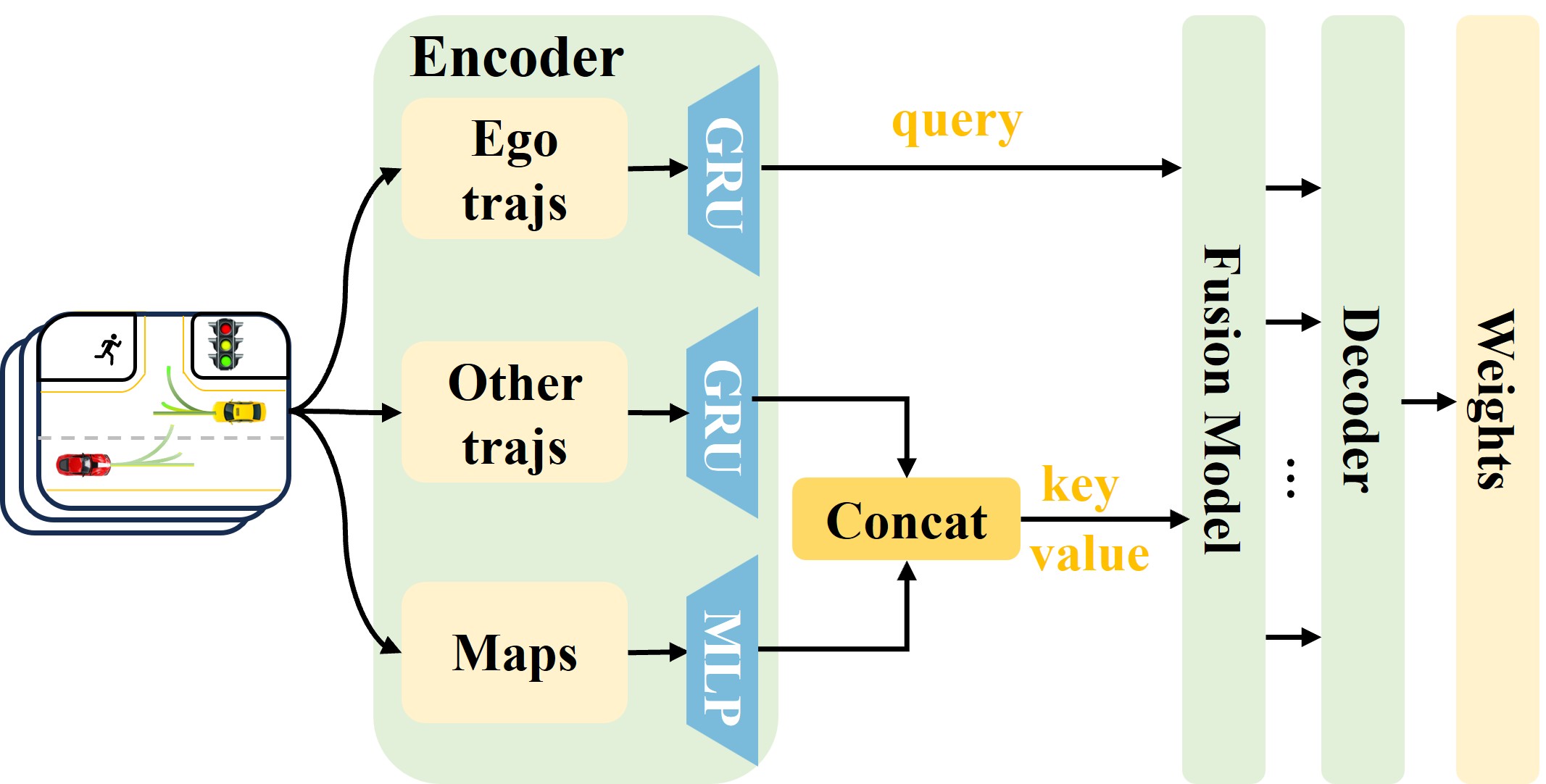}
      \caption{Architecture of the scene-aware neural network.}
      \label{Network}
\end{figure}

\textbf{Cost Function}: The cost function depends jointly on the states and controls of multiple agents, making their interactions difficult to specify explicitly. Following~\cite{wulfmeier2015maximum}, we use a neural network to model these latent couplings, as shown in \textit{Line 11 of Algorithm~\ref{Alg2}}. The cost function is defined as follows.
\begin{equation}
    \boldsymbol{\omega}^i=\phi_{\theta^i}(\zeta,c),
\end{equation}
where $c$ denotes an optional input representing scene information, such as road boundaries, traffic lights, etc. When no such information is available, the input is padded with zeros. The architecture of the network $\phi_{\theta^i}$ is shown in Fig.~\ref{Network} with the following modules 
\begin{itemize}
\item \textit{Encoder:} Agent trajectories are encoded via a 
GRU network, and optional vectorized map features are embedded 
through an MLP following~\cite{vectornet}.
\item \textit{Fusion Module:} A cross-attention mechanism 
integrates ego-agent, neighboring-agent, and map 
representations to produce interaction-aware feature vectors.
\item \textit{Decoder:} A fully connected layer maps the fused 
representation to the cost weight vector 
$\boldsymbol{\omega}^i$.
\end{itemize}

\textbf{Trajectory probability}: The likelihood of an expert trajectory is modeled by a Boltzmann distribution, where each trajectory is weighted exponentially by its reward. This distribution is the maximum-entropy distribution that matches the expert’s expected feature statistics~\cite{ziebart2008maximum}. Thus, the objective in Eq.~\eqref{eq:inv_obj} can be equivalently written as maximizing
\begin{equation}
  P_i(\zeta |\boldsymbol{\omega} ^i)\approx \frac{e^{-R_{\boldsymbol{\omega} ^i}(\zeta )}}{\sum{_{j=1}^{M}e^{-R_{\boldsymbol{\omega} ^i}(\tilde{\zeta} ^j)}}},
\label{Ptheta}
\end{equation}
where $\tilde{\zeta}^j$ is a rollout trajectory with the same initial state as expert trajectory $\zeta$, and the cost is the weighted sum of the individual cost primitives over all time steps, denoted as
\begin{equation}
\begin{aligned}
R_{\boldsymbol{\omega}^i}(\zeta) 
&= \sum\limits_{t=1}^T c_t^i = \boldsymbol{\omega}^{i,\top} \mathbf{f}^i_\zeta = \boldsymbol{\omega}^{i,\top} \sum\limits_{t=1}^T \mathbf{f}_t^i(\mathbf{x}_t, \mathbf{u}_t),
\end{aligned}
\label{TrajR}
\end{equation}
where $\mathbf{f}_t^i(\cdot)$ denotes predefined cost primitive function for agent $i$, which captures essential cost factors such as distance or energy usage. These primitives are linearly combined with weights $\boldsymbol{\omega}^i$ to form the total cost function. We use this equation in \textit{Lines 2 and 6 of Algorithm~\ref{Alg2}}.
It is worth noting that although the combination is linear, the primitives themselves can encode nonlinear effects, enabling the model to approximate complex cost landscapes. Details are provided in Section~\ref{primitive_funtion}.

\textbf{Loss Function}: The total training objective comprises the maximum-entropy IRL loss and the rule penalty term, as shown in \textit{Line 17 of Algorithm~\ref{Alg2}}, formulated as follows
\begin{equation}
    \mathcal{L} _{\mathrm{total}}^{i}=\mathcal{L} _{\mathrm{MaxEnt}}^{i}+\nu_{\mathrm{rule}}\cdot \mathcal{L} _{\mathrm{rule}}^{i},
\label{17}
\end{equation}
where \(\nu_{\mathrm{rule}}\) is a weight that balances the two loss terms. 
\(\mathcal{L}_{\mathrm{MaxEnt}}^i\) denotes the MaxEnt loss computed in Line 14 of Algorithm~\ref{Alg2}, and is defined as
\begin{equation}
    \mathcal{L}_{\mathrm{MaxEnt}}^{i} = \sum\limits_{\zeta \in \mathcal{D}} \left[ R_{\boldsymbol{\omega}^i}(\zeta) \! + \!\log \sum\limits_{j=1}^{M} \exp \left( -R_{\omega ^i}(\tilde{\zeta}^j) \right) \right].
\end{equation}

Additionally, $\mathcal{L}_{\mathrm{rule}}^{i}$ denotes the penalty associated with physical rules, computed in \textit{Line 15 of Algorithm~\ref{Alg2}}. In multi-agent decision-making, cost functions learned from expert data alone may yield policies that violate task constraints or are physically infeasible. To mitigate this issue, we augment the training objective with rule-based loss terms derived from predefined task rules. Let $\mathbf{f}_t^{i,\mathrm{rule}} \subseteq \mathbf{f}_t^i$ denote the subset of cost primitives corresponding to physical rules, such as the collision-avoidance term $f_2^i$ and the lane-boundary term $f_3^i$ in Section~\ref{primitive_funtion}. The rule penalty is defined as:
\begin{equation}
    \mathcal{L}_{\mathrm{rule}}^i
    = \mathbb{E}_{\zeta \sim \mathcal{D}}
    \left[
        \sum_{t=1}^{T}
        \left\|
            \mathbf{f}_t^{i,\mathrm{rule}}(\mathbf{x}_t, \mathbf{u}_t)
        \right\|_2
    \right].
\end{equation}
\section{Experiments}
In this section, we evaluate the proposed method in collision avoidance and 
high-speed ramp merging scenarios, comparing against three 
baselines with multiple metrics and further validating on 
hardware.
\subsection{Experimental Setup}
\subsubsection{Baselines}
We compare against three baselines:
\textbf{(i) KKT-based}: the open-loop Nash solver of Peters et al.~\cite{peters2023online}, included to compare open-loop and feedback strategies;
\textbf{(ii) MPC-based}: the prediction-then-planning framework of Lubars et al.~\cite{lubars2021combining}, which models other agents with constant-velocity predictions;
\textbf{(iii) KLGame}: the KL-regularized game formulation of Lidard et al.~\cite{lidard2024blending}, which uses a simple MLP for cost inference and lacks adaptive data--rule weighting, highlighting the benefits of our scene-aware network and weighting mechanism.
\subsubsection{Scenarios}
We evaluate our approach in two representative multi-agent interactive scenarios:
\textbf{(i) Collision Avoidance Multi-agent Planning (CAMP)}: a coordinated position-exchange task in which agents start from opposing locations and must avoid collisions while minimizing travel distance;
\textbf{(ii) Ramp Merging (RM)}: a high-speed five-agent merging scenario in which vehicles negotiate limited space under traffic rules and safety constraints.
\subsubsection{Dynamics and cost primitive functions} \label{primitive_funtion}
In all scenarios, each agent $i$ follows a 4D unicycle dynamics model, given in Eq.~\eqref{4D} with a time discretization step $\Delta t$.
\begin{equation}
    \begin{bmatrix}
p^i_{x,t+1} \\
p^i_{y,t+1} \\
\beta^i_{t+1} \\
v^i_{t+1}
\end{bmatrix}
=
\begin{bmatrix}
p^i_{x,t} \\
p^i_{y,t} \\
\beta^i_t \\
v^i_t
\end{bmatrix}
+ \Delta t
\begin{bmatrix}
v_t^i \cos(\beta_t^i) \\
v_t^i \sin(\beta_t^i) \\
\Omega_t^i \\
a_t^i
\end{bmatrix},
\label{4D}
\end{equation}
where ${{\mathbf{x}}_t} = [p_{x,t}^1,p_{y,t}^1,\beta _t^1,v_t^1, \ldots p_{x,t}^N,p_{y,t}^N,\beta _t^N,v_t^N]$ is the joint state collecting the position, orientation, and velocity of all agents, and $\mathbf{u}_t^i = [\Omega _t^i,a_t^i] \in {\mathbb{R}^2}$ is the control input of the agent $i \in [N]$, including the yaw rate $\Omega _t^i$ and the longitudinal acceleration $a_t^i$. Also, we discretize the time $\Delta t=0.1\,\mathrm{s}$.
 
Regarding the design of the cost function, we can represent it as $c_t^i = {\boldsymbol{\omega}^{i,\top}} \mathbf{f}_t^i$. The cost primitive function $ \mathbf{f}^i$ is given by 
\begin{equation}
\mathbf{f}^i_t=\left[ \mathrm{f}_{1}^{i},\mathrm{f}_{2}^{i},\mathrm{f}_{3}^{i},\mathrm{f}_{4}^{i} \right]\left\{ \begin{gathered}
  \mathrm{f}_1^i = \left\| {p_t^i - p_{goal}^i} \right\|_2^2 \\
  \mathrm{f}_2^i =  - \sum\nolimits_{j\ne i} {\log } \left( {\left\| {{p_t^i} - {p_t^j}} \right\|_2^2} \right) \\
  \mathrm{f}_3^i = \mathbf{1}\{d_l(p_t^i) > d_{lane}\}
(d_l(p_t^i) - d_{lane})^2 \\
  \mathrm{f}_4^i = \sum_{j=1}^N \mathbf{u}_t^{j\top} R_t^{ij} \mathbf{u}_t^j
\end{gathered}  \right.
\end{equation}
where ${{\boldsymbol{\omega}}^i} = {[\omega_l^i]_{l \in [4]}}$, 
$\omega_l^i \in {\mathbb{R}_+}$, and $p_t^i=[p_{x,t}^i,p_{y,t}^i]$ 
(the time index $t$ of each $\mathrm{f}_k^i$ is omitted for brevity). 
Here $\mathrm{f}_1^i$ penalizes the squared distance to a fixed goal, 
$\mathrm{f}_2^i$ enforces inter-agent collision avoidance via a 
log-barrier, $\mathrm{f}_3^i$ penalizes lane-boundary violations 
(used in scenario \textbf{RM}, where $d_{lane}$ is the lane 
half-width), and $\mathrm{f}_4^i$ regularizes the control effort.
\subsubsection{Implementation Details}
The network $\phi_\theta^i$ is trained with Adam ($\alpha=1\times10^{-3}$). The GRU encoder 
uses a hidden dimension of $d_h=128$ over a history window 
of $H=10$ steps ($1.0\,\mathrm{s}$ at $\Delta t=0.1\,\mathrm{s}$); the 
attention fusion module has $n_{\text{head}}=4$ heads with 
$d_k=64$; the decoder is a two-layer MLP with hidden 
dimension 64 and Softplus output to enforce 
$\boldsymbol{\omega}^i \succ 0$. Training uses $K=500$ expert trajectories 
with $M=30$ rollouts each. Key hyperparameters are fixed as 
$\nu_{\text{rule}}=0.1$, $\sigma=1.0$, 
$\lambda_{\min}^i=0.5$, $\lambda_{\max}^i=5.0$ across all 
scenarios. All experiments run on four NVIDIA RTX 3090 GPUs.
\subsubsection{Metrics}\label{Metrics}
To evaluate the level of consistency between the ground-truth data and the planned trajectories, the following metrics are defined: number of collisions $\mathbf{Coll}$, cost function error $D_{cos}$, trajectory error $D_{tra}$, and parameter error $D_{par}$. The $\mathbf{Coll}$ denotes the number of collision events between agents, and $D_{cos}$, $D_{par}$, $D_{tra}$ are calculated as follows:
\begin{subequations}
\setlength{\jot}{-2pt}
    \begin{equation}
        {D_{\cos }}({\boldsymbol{\omega} _{true}},{\boldsymbol{\omega} _{est}}) = \frac{1}{N}\sum\limits_{i \in [N]} {\left\| {J_{true}^i - J_{est}^i} \right\|_2},
    \end{equation}
    \begin{equation}
         {D_{tra}}({{\boldsymbol{\omega }}_{{\text{true}}}},{{\boldsymbol{\omega }}_{{\text{est}}}}) = \frac{1}{{NT}}\sum\limits_{t \in [T]} {\sum\limits_{i \in [N]} {{{\left\| {p_{t,est}^i - p_{t,true}^i} \right\|}_2}} },
    \end{equation}
    \begin{equation}
        {D_{par }}({\boldsymbol{\omega} _{{\text{true}}}},{\boldsymbol{\omega} _{{\text{est}}}}) = 1 - \frac{1}{N}\sum\limits_{i \in [N]} \frac{{{\boldsymbol\omega }}_{{\text{true}}}^{i,\top}{\boldsymbol{\omega }}_{est}^i}{{{{\left\| {{\boldsymbol{\omega }}_{{\text{true}}}^i} \right\|}_2}{{\left\| {{\boldsymbol{\omega }}_{est}^i} \right\|}_2}}},
    \end{equation}
\end{subequations}
where $\boldsymbol{\omega}_{\text{true}}$ and $\boldsymbol{\omega}_{\text{est}}$ represent the ground-truth and estimated cost parameters, respectively. $p^i_{t,\text{true}}$ and $p^i_{t,\text{est}}$ denote the true and reconstructed positions of agent $i$ at time step $t$, obtained from the NE solution using $\boldsymbol{\omega}^i_{\text{est}}$. The corresponding true and estimated costs are $J^i_{\text{true}}$ and $J^i_{\text{est}}$. 
\subsection{Simulation Results}
\subsubsection{Effectiveness of our method}
\begin{figure}[htbp]
\centering
    \begin{minipage}[b]{0.44\linewidth}
        \centering
        \includegraphics[width=\linewidth, height=0.9\linewidth]{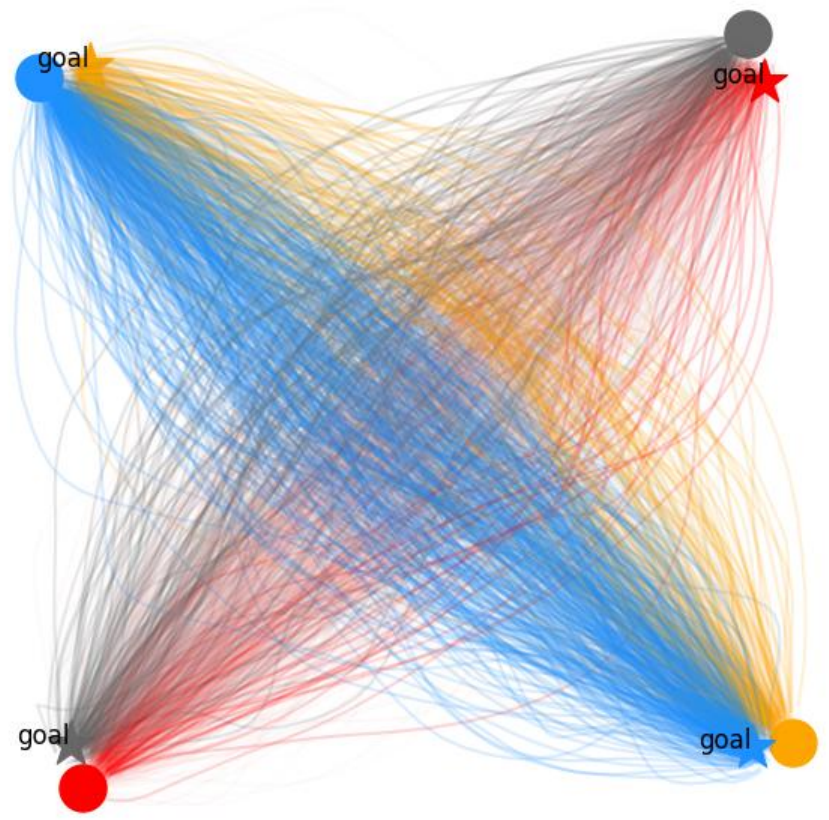}
        \centerline{(a)}
    \end{minipage}
    \begin{minipage}[b]{0.44\linewidth}
        \centering
        \includegraphics[width=\linewidth, height=0.9\linewidth]{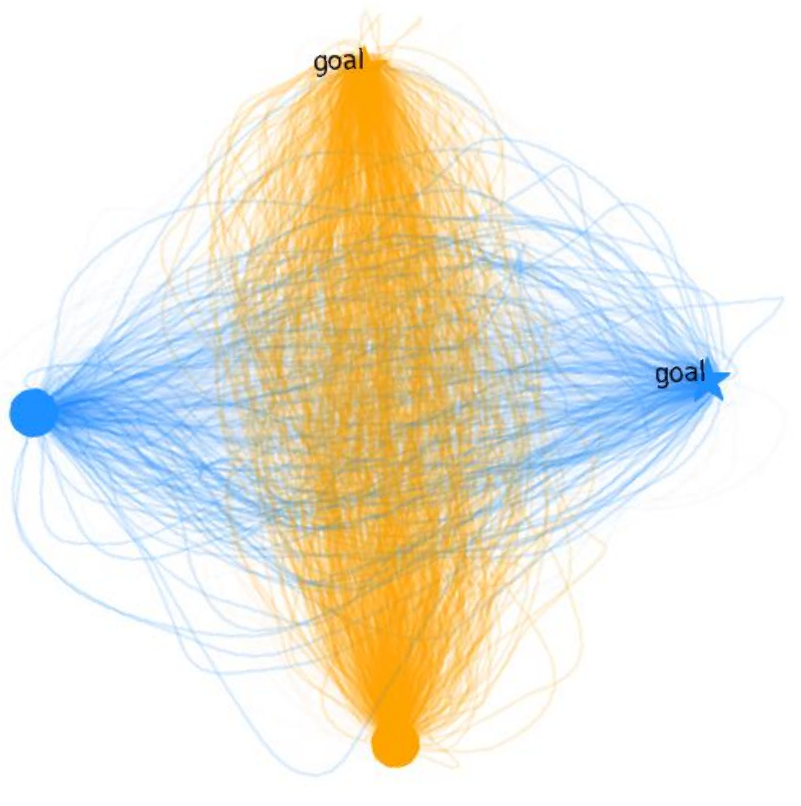}
        \centerline{(b)}
    \end{minipage}
\caption{\textbf{CAMP}: (a) Four agents perform pairwise position exchange. (b) Two agents perform the same exchange task}
\label{scene1_results}
\end{figure}

\begin{figure}[htbp]
      \centering
      \includegraphics[width=\linewidth, height=0.4\linewidth]{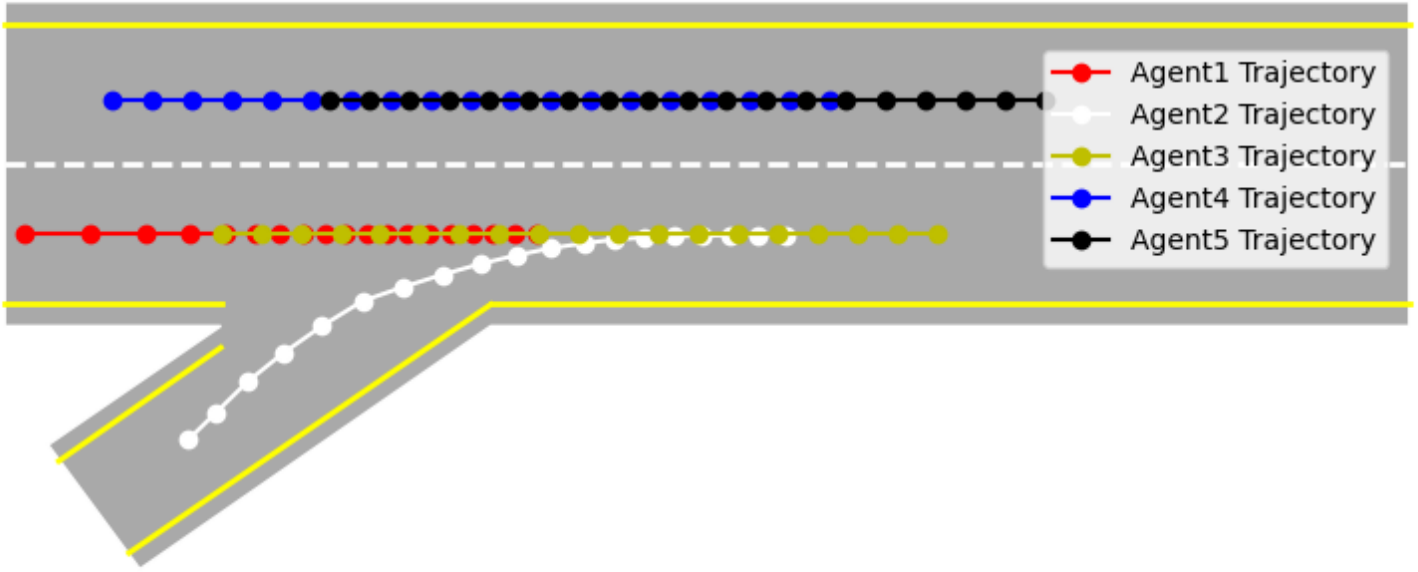}
      \caption{\textbf{RM}: Agent2 completes a collision-free merge. Agent4 and 5 are excluded from the interaction due to lane separation but included for metric evaluation.}
      \label{scene2_results}
\end{figure}

The results for \textbf{CAMP} are shown in Fig.~\ref{scene1_results}. Across 500 trials with the learned strategy, all agents successfully reached their goals without collisions. The results for \textbf{RM} are presented in Fig.~\ref{scene2_results}, where five self-driving vehicles perform interactive planning in a congested merging scenario. By jointly considering collision avoidance with other agents and lane constraints, the proposed method generates feasible trajectories that successfully accomplish the ramp merging task.

\begin{figure}[htbp]
      \centering
      \includegraphics[width=\linewidth, height=0.9\linewidth]{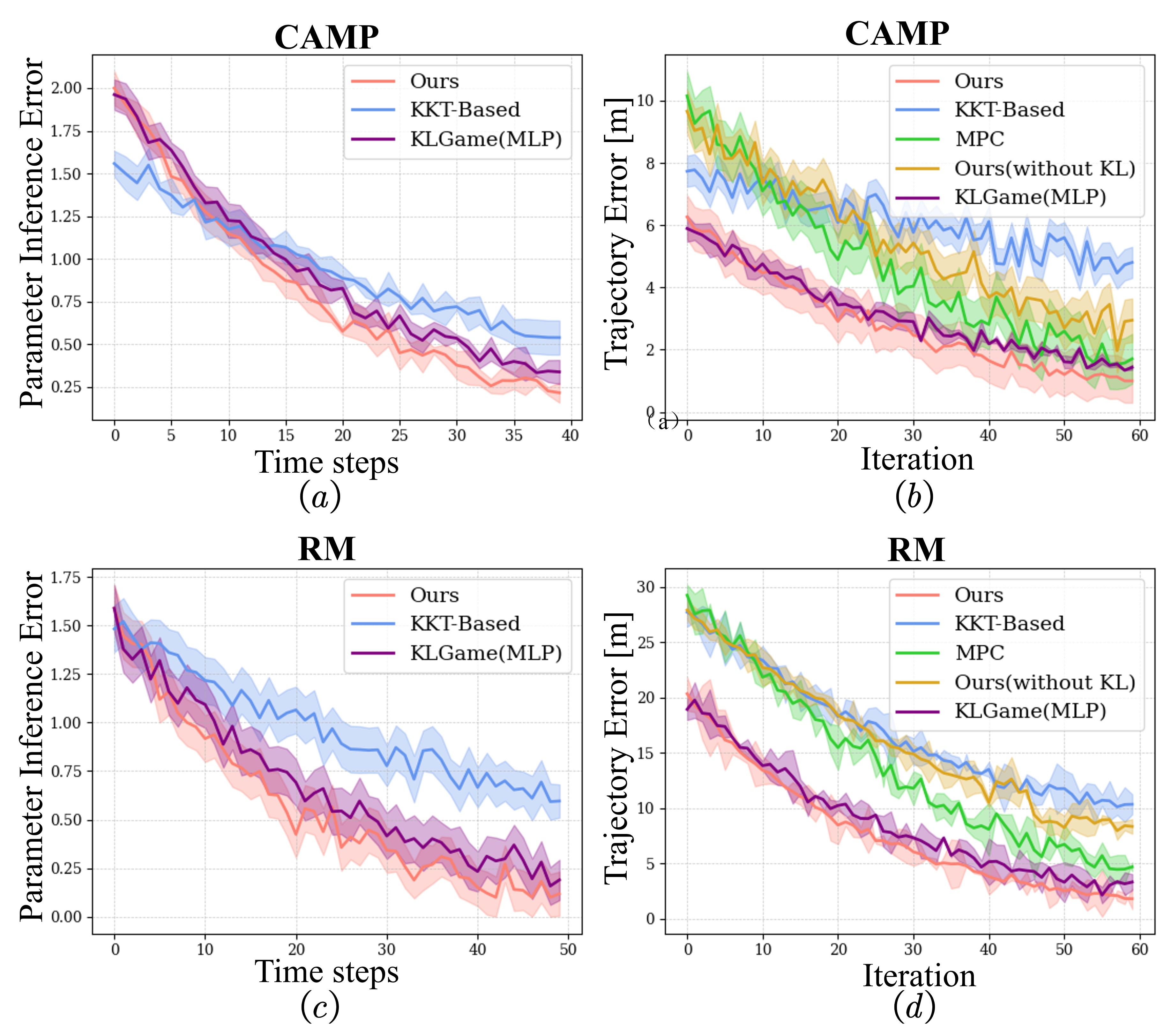}
      \caption{Comparison of trajectory and parameter errors.}
      \label{Error}
\end{figure}

To further evaluate the role of the KL term, we removed it after estimating the cost weights and solved for the corresponding NE solution. As shown in Fig.~\ref{Error}(b)(d), the KL-regularized formulation achieves smaller errors with respect to the expert trajectories under the same number of iterations. In contrast, removing the KL term leads to slower convergence, indicating that KL regularization improves strategy learning from expert data and reduces trajectory error.

To assess the contribution of the proposed network, we disabled its weight-fitting module and instead recovered the cost weights $\boldsymbol{\omega}$ via gradient-based estimation following~\cite{li2023cost}, while keeping all other settings unchanged. Over 50 trials, this ablated variant underperforms the full model in both parameter estimation accuracy and trajectory error, as reported in Table~\ref{table1}. These results confirm that the proposed network more effectively captures scene-aware map information and the implicit relationships among cost terms.
\subsubsection{Ablation Study and Comparison with Baselines}

We conducted 50 independent trials under identical settings. 
To disentangle the two key contributions, we introduce 
\textbf{Ours(fixed~$\lambda$)}, which retains the scene-aware 
network but replaces $\lambda^i(\mathbf{x}_t)$ with a fixed 
scalar. The gap between \textbf{Ours} and 
\textbf{Ours(fixed~$\lambda$)} thus isolates the benefit of 
adaptive regularization, while the gap between \textbf{Ours} 
and \textbf{Ours(no net)} isolates the benefit of the 
scene-aware network.

In the \textbf{CAMP} scenario, all methods perform comparably 
in the two-agent task: \textbf{Ours(fixed~$\lambda$)} nearly 
matches \textbf{Ours} ($D_{par}$: 0.19 vs.\ 0.20), confirming 
that adaptive regularization provides negligible benefit when 
interactions are simple. As complexity grows to four agents, 
the gap widens: \textbf{Ours(fixed~$\lambda$)} reduces 
$D_{par}$ from $1.18$ (KLGame) to $1.49$, while 
\textbf{Ours} further reduces it to $1.18$, showing that the 
scene-aware network and adaptive $\lambda^i(\mathbf{x}_t)$ 
each contribute independently. In the 20-agent setting, the 
benefit of adaptive regularization becomes dominant: 
\textbf{Ours} reduces collisions to 4 vs.\ 6 for 
\textbf{Ours(fixed~$\lambda$)} and 7 for \textbf{KLGame}, 
and achieves the lowest $D_{par}$ (3.82) and $D_{tra}$ 
(18.63), demonstrating that state-dependent regularization 
becomes increasingly critical as interaction density grows.

In the \textbf{RM} scenario, \textbf{Ours(fixed~$\lambda$)} 
reduces collisions from 2 (KLGame) to 1, but only the 
full model with adaptive $\lambda^i(\mathbf{x}_t)$ achieves 
zero collisions, confirming that tightening reference-policy 
adherence near the high-risk merge zone is the key mechanism 
for safety. The scene-aware network provides an additional 
reduction in $D_{par}$ from $1.03$ to $0.88$ and in 
$D_{tra}$ from $12.31$ to $11.03$ relative to 
\textbf{Ours(fixed~$\lambda$)}, confirming that jointly 
encoding map context captures 
interaction structure that fixed-coefficient methods cannot 
represent.
\begin{table}[htbp]
    \renewcommand{\arraystretch}{0.9}
    \centering
    \caption{Quantitative results are reported as mean $\pm$ standard 
    deviation over 50 trials.}
    \resizebox{\columnwidth}{!}{
    \begin{tabular}{c|c|c|c|c|c}
        \toprule
        \textbf{Scenarios} & \textbf{Methods}
        & \textbf{Cost err $D_{\mathrm{cos}}$} & \textbf{Coll}
        & \textbf{Param err $D_{\mathrm{par}}$} & \textbf{Traj err $D_{\mathrm{tra}}$} \\
        \midrule
        & Ours
            & 3.82$\pm$1.41 & \textbf{0} & 0.20$\pm$0.22 & 0.13$\pm$0.31 \\
        CAMP
        & Ours(fixed $\lambda$)
            & 3.79$\pm$1.30 & \textbf{0} & 0.19$\pm$0.20 & 0.13$\pm$0.28 \\
        (2 Agents)
        & Ours(no net)
            & 3.73$\pm$0.91 & 2          & \textbf{0.11$\pm$0.12} & 0.12$\pm$0.23 \\
        & KKT-Based~\cite{peters2023online}
            & 3.88$\pm$0.84 & \textbf{0} & 0.12$\pm$0.15 & 0.12$\pm$0.20 \\
        & MPC~\cite{lubars2021combining}
            & $n/a$         & \textbf{0} & $n/a$         & \textbf{0.12$\pm$0.15} \\
        & KLGame~\cite{lidard2024blending}
            & \textbf{3.72$\pm$1.20} & \textbf{0} & 0.18$\pm$0.22 & 0.15$\pm$0.11 \\
        \midrule
        & Ours
            & \textbf{5.82$\pm$1.20} & \textbf{2} & \textbf{1.18$\pm$0.35} & \textbf{3.21$\pm$1.82} \\
        CAMP
        & Ours(fixed $\lambda$)
            & 5.84$\pm$1.38 & 3          & 1.49$\pm$0.41 & 3.86$\pm$1.94 \\
        (4 Agents)
        & Ours(no net)
            & 6.91$\pm$1.12 & \textbf{2} & 3.13$\pm$1.21 & 5.32$\pm$1.22 \\
        & KKT-Based~\cite{peters2023online}
            & 6.60$\pm$2.11 & 6          & 1.22$\pm$0.53 & 7.25$\pm$0.52 \\
        & MPC~\cite{lubars2021combining}
            & $n/a$         & \textbf{2} & $n/a$         & 5.22$\pm$0.86 \\
        & KLGame~\cite{lidard2024blending}
            & 5.83$\pm$2.11 & 3          & 1.18$\pm$0.47 & 3.27$\pm$1.21 \\
        \midrule
        & Ours
            & \textbf{18.52$\pm$3.21} & \textbf{4} & \textbf{3.82$\pm$0.85} & \textbf{18.63$\pm$4.21} \\
        CAMP
        & Ours(fixed $\lambda$)
            & 19.13$\pm$3.48 & 6         & 4.43$\pm$0.90 & 20.17$\pm$3.84 \\
        (20 Agents)
        & Ours(no net)
            & 21.34$\pm$3.82 & 8         & 6.51$\pm$1.43 & 24.71$\pm$3.92 \\
        & KKT-Based~\cite{peters2023online}
            & 25.81$\pm$5.14 & 15        & 5.63$\pm$1.21 & 32.44$\pm$5.83 \\
        & MPC~\cite{lubars2021combining}
            & $n/a$          & 12        & $n/a$         & 28.92$\pm$6.24 \\
        & KLGame~\cite{lidard2024blending}
            & 19.24$\pm$4.13 & 7         & 4.15$\pm$0.92 & 21.35$\pm$3.61 \\
        \midrule
        & Ours
            & \textbf{12.21$\pm$2.08} & \textbf{0} & \textbf{0.88$\pm$0.22} & \textbf{11.03$\pm$2.50} \\
        RM
        & Ours(fixed $\lambda$)
            & 12.48$\pm$2.24 & 1         & 1.03$\pm$0.31 & 12.31$\pm$1.68 \\
        (5 Agents)
        & Ours(no net)
            & 12.53$\pm$2.36 & 2         & 1.24$\pm$0.47 & 13.37$\pm$1.52 \\
        & KKT-Based~\cite{peters2023online}
            & 15.62$\pm$3.11 & 3         & 1.38$\pm$0.32 & 16.09$\pm$2.41 \\
        & MPC~\cite{lubars2021combining}
            & $n/a$          & 6         & $n/a$         & 15.26$\pm$4.31 \\
        & KLGame~\cite{lidard2024blending}
            & 12.73$\pm$2.80 & 2         & 0.92$\pm$0.25 & 13.20$\pm$1.14  \\
        \bottomrule
    \end{tabular}
    }
    \label{table1}
\end{table}
\begin{table}[htbp]
    \renewcommand{\arraystretch}{0.9}
    \centering
    \caption{Sensitivity of key metrics to $\nu_{\mathrm{rule}}$ in the RM scenario (5 agents, 50 trials).}
    \label{tab:sensitivity}
    \resizebox{0.85\columnwidth}{!}{
    \begin{tabular}{ccccc}
        \toprule
        $\nu_{\mathrm{rule}}$ 
        & Param err $D_{\mathrm{par}}$ 
        & Traj err $D_{\mathrm{tra}}$ 
        & Coll 
        & Interpretation \\
        \midrule
        0    
        & 1.14$\pm$0.28 
        & 12.62$\pm$2.83 
        & 2 
        & Pure MaxEnt IRL \\
        0.01 
        & 0.93$\pm$0.24 
        & 11.28$\pm$2.56 
        & \textbf{0} 
        & Weak constraint \\
        0.1  
        & \textbf{0.88$\pm$0.22} 
        & \textbf{11.03$\pm$2.50} 
        & \textbf{0} 
        & \textbf{Default} \\
        1.0  
        & 0.96$\pm$0.25 
        & 11.47$\pm$2.61 
        & \textbf{0} 
        & Strong constraint \\
        10.0 
        & 1.29$\pm$0.31 
        & 13.74$\pm$2.95 
        & \textbf{0} 
        & Near rule-only \\
        \bottomrule
    \end{tabular}
    }
\end{table}
\vspace{-5mm}
\subsection{Sensitivity Analysis of \texorpdfstring{$\nu_{\mathrm{rule}}$}{nu\_rule}}
To evaluate the sensitivity of the method to the hyperparameter $\nu_{\mathrm{rule}}$ in Eq.~\eqref{17}, we perform an ablation study in the RM scenario by varying $\nu_{\mathrm{rule}} \in \{0, 0.01, 0.1, 1.0, 10.0\}$ while keeping all other settings fixed. 

As shown in Table~\ref{tab:sensitivity}, the method is stable over a broad range $\nu_{\mathrm{rule}} \in [0.01,1.0]$, indicating low sensitivity to the exact choice of this hyperparameter. When $\nu_{\mathrm{rule}}=0$, removing the rule loss increases the collision count, showing that the physical constraint term contributes directly to safety. In contrast, an overly large value ($\nu_{\mathrm{rule}}=10.0$) over-constrains the learned cost and degrades trajectory accuracy. These two extremes correspond to interpretable failure modes, confirming that $\nu_{\mathrm{rule}}$ serves as a meaningful trade-off between data-driven imitation and rule enforcement.
\begin{figure}[htbp]
  \centering
  \begin{minipage}[t]{\linewidth}
    \centering
    \includegraphics[width=0.97\linewidth]{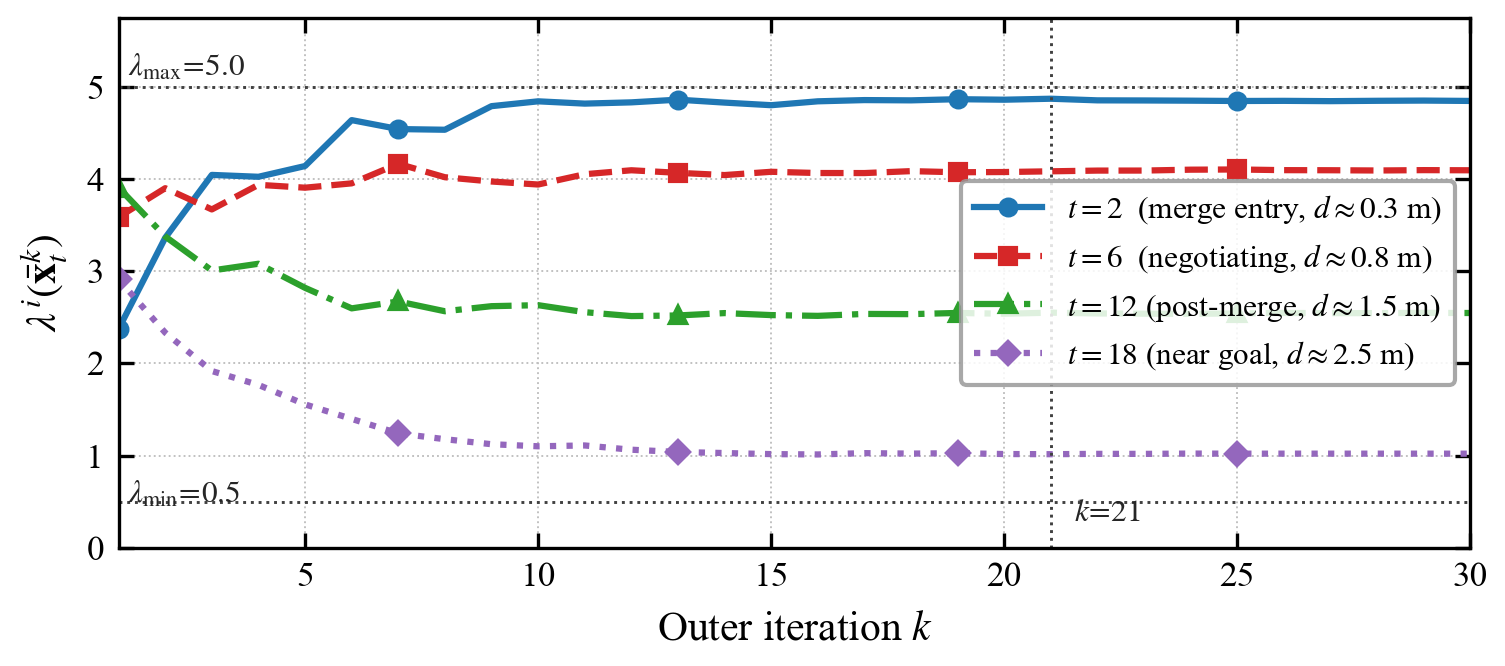}
    \centerline{\small (a) Evolution of $\lambda^i(\hat{\mathbf{x}}_t^k)$ across outer iterations.}
  \end{minipage}
  \begin{minipage}[t]{\linewidth}
    \centering
    \includegraphics[width=0.97\linewidth]{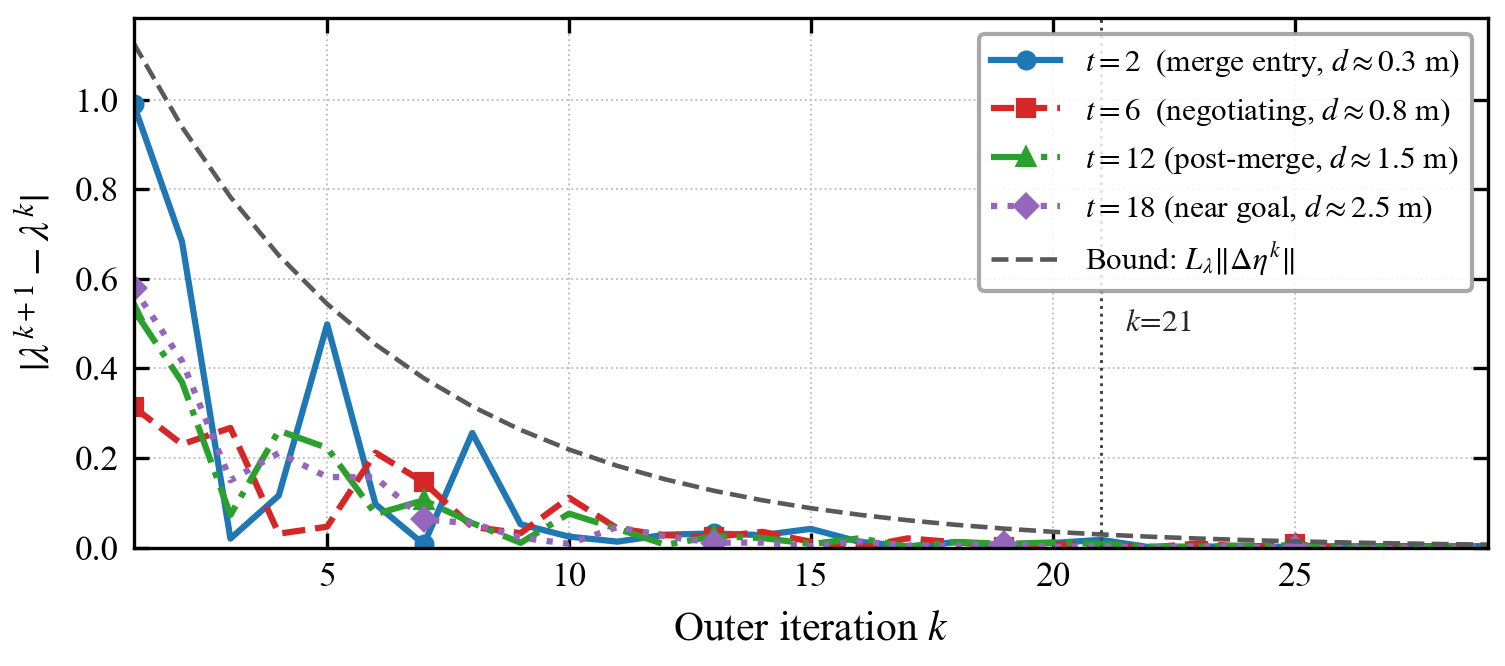}
      \centerline{\small (b) Per-iteration perturbation $|\lambda^{k+1}-\lambda^k|$ across outer iterations.}
  \end{minipage}
  \caption{State-dependent $\lambda^i$ convergence in the \textbf{RM} (5 agents) scenario.}
  \label{fig:lambda_conv}
\end{figure}
\subsection{Convergence of State-Dependent \texorpdfstring{$\lambda^i$}{lambda\^{}i}}
To illustrate the behavior of the adaptive regularization, we trace
$\lambda^i(\hat{\mathbf{x}}_t^k)$ at four representative planning steps
(each corresponding to $\Delta t = 0.1$\,s) in the \textbf{RM} scenario across outer iterations of \textit{Algorithm~\ref{Ag1}}.
As shown in Fig.~\ref{fig:lambda_conv}(a), the converged values range from
$\lambda^*\approx4.85$ near the merge zone ($d\approx0.3$~m) to
$\lambda^*\approx1.02$ near the goal ($d\approx2.5$~m), confirming that
the proposed mechanism assigns stronger regularization in regions of high
interaction density while granting greater optimization freedom elsewhere.
Fig.~\ref{fig:lambda_conv}(b) further shows that the per-iteration
perturbation $|\lambda^{k+1}-\lambda^k|$ remains bounded and decays
rapidly, consistent with Proposition~\ref{prop:lambda_regular}.
\vspace{-3mm}
\subsection{Hardware Experiments}
To validate our framework in real-world scenarios, we conducted hardware experiments involving two TurtleBot4s and a human participant. Each robot used an Intel RealSense T265 camera\footnote{https://www.intelrealsense.com/tracking-camera-t265/} for $100\,\mathrm{Hz}$ visual-inertial pose estimation. We evaluated two tasks: (i) position exchange, (ii) trajectory crossing. Agents navigated to private goal positions without accessing others' objectives. Strategies were trained offline using joystick-collected human demonstrations and deployed online. Results are shown in Fig.~\ref{Hardware}; video demonstrations and implementation details are in the supplementary material. All computations ran on an Intel NUC (Core i7-1260P). 
With warm-starting and a maximum of $k_{\max}=15$ 
iterations, the average per-step planning time was 
$35\,\mathrm{ms}$ ($\approx$28\,Hz), satisfying the $10\,\mathrm{Hz}$
real-time requirement imposed by $\Delta t=0.1\,\mathrm{s}$.
\begin{figure}[htbp]
      \centering
      \includegraphics[width=\linewidth, height=0.6\linewidth]{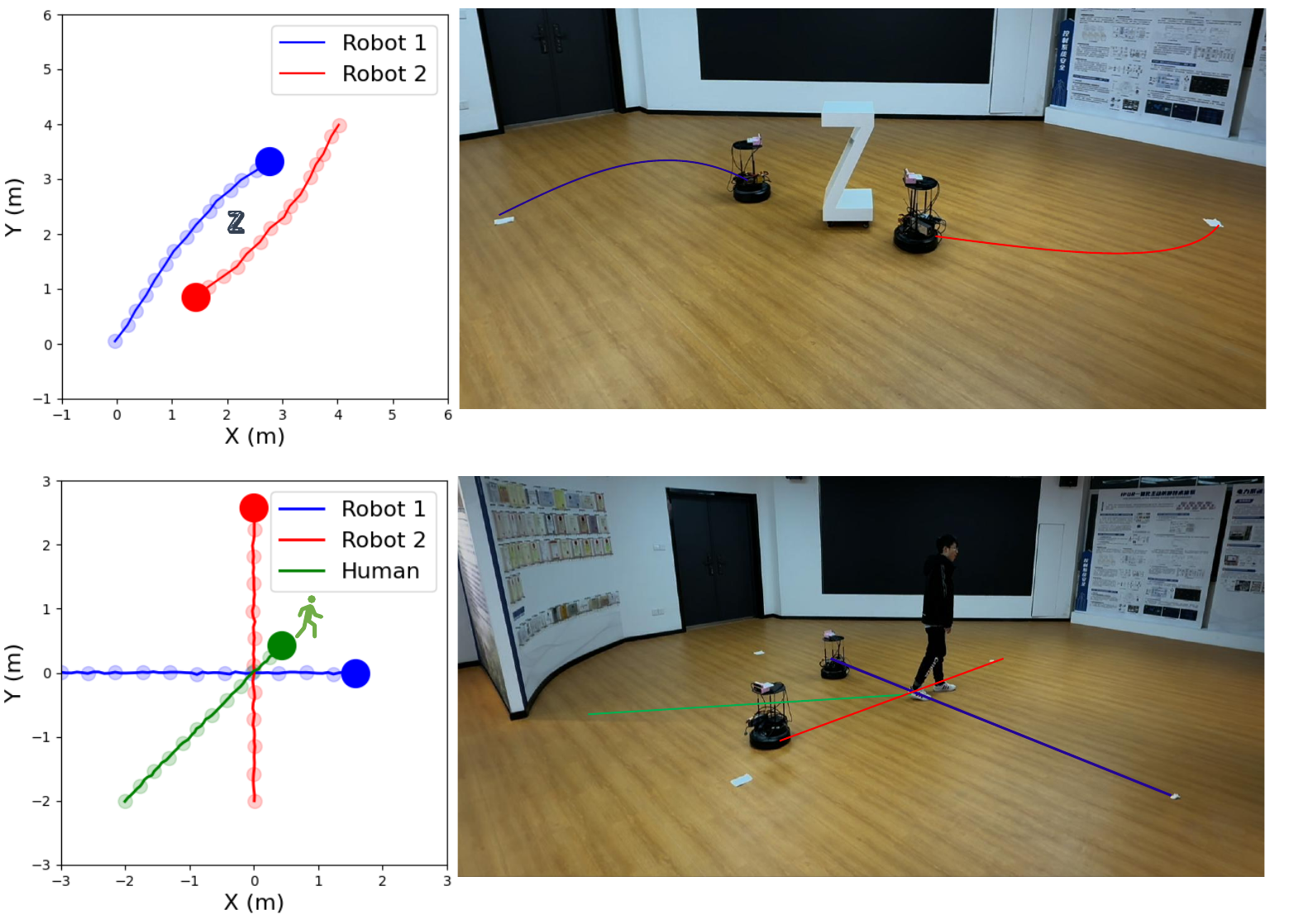}
      \caption{Trajectories executed using learned strategies in position exchange (top) and trajectory crossing (bottom).}
      \label{Hardware}
\end{figure}
\vspace{-3mm}
\section{Conclusion}
We proposed a forward--inverse dynamic game framework for multi-agent
trajectory planning that couples scene-aware cost inference with a
KL-regularized game solver.
A state-dependent regularization weight $\lambda^i(\mathbf{x}_t)$
automatically adapts the policy distribution to interaction density,
with theoretical guarantees on per-iteration well-posedness and the
Lipschitz continuity of the adaptive regularization weight, as established
in Propositions~\ref{prop:lambda_wellposed} and~\ref{prop:lambda_regular}.
Experiments confirm competitive
parameter recovery and trajectory accuracy. Two limitations should be noted. First, expert demonstrations are currently obtained from forward simulation rather than naturalistic datasets such as INTERACTION~\cite{zhan2019interaction}. Second, future work will infer agent intentions directly from raw sensory inputs, toward a fully end-to-end multi-agent interaction framework.
\appendix

\section{Proof of Lemma~\ref{lemma:variational} and Theorem~\ref{th1}}
\label{appendix:proof}
\subsection{Dynamic Programming Formulation}

For each agent $i$, let $\mathbf{x}\in\mathcal{X}$ denote the joint system state,
$\mathbf{u}^i\in\mathcal{U}^i$ the control action, $\pi^i(\cdot|\mathbf{x})$ its mixed
strategy, and $\tilde{\pi}^{\,i}(\cdot|\mathbf{x})$ the reference policy.
Define the $\mathcal{Q}$-value function as
\begin{equation}
\mathcal{Q}^{i}(\mathbf{x},\mathbf{u})=c^{i}(\mathbf{x},\mathbf{u})+V^{i}(f(\mathbf{x},\mathbf{u})),
\end{equation}
where $c^{i}$ is the stage cost and $f(\cdot)$ is the system dynamics.
Applying the dynamic programming principle yields the Bellman equation
\begin{equation}
V^i(\mathbf{x})\!
=\!\inf_{\pi^i}\Bigl\{
\mathbb{E}^{\pi}\!\bigl[\mathcal{Q}^i(\mathbf{x},\mathbf{u})\bigr]
\!+\!\lambda^i D_{\mathrm{KL}}\!\bigl(\pi^i(\cdot|\mathbf{x})\,\|\,\tilde{\pi}^{\,i}(\cdot|\mathbf{x})\bigr)\!
\Bigr\},
\label{eq:bellman_app}
\end{equation}
where $V^i(\mathbf{x})=\inf_{\pi^i}J^i(\pi^i)$ is the optimal value function, and the KL term
penalizes deviations of the current policy from the reference distribution at state $\mathbf{x}$. For convenience in the subsequent derivations, we omit the time index $t$ and the agent index $i$ unless necessary.
\subsection{Proof of Lemma~\ref{lemma:variational}}
\begin{proof}
\textbf{Step 1: Deriving the optimal density via a variational argument.}

To characterize the optimal policy, we introduce the Radon--Nikodym derivative of $\pi(\cdot\mid\mathbf{x})$ with respect to the reference measure $\tilde{\pi}(\cdot\mid\mathbf{x})$, defined as
\begin{equation}
p(\mathbf{u}):=\frac{\mathrm{d}\pi}{\mathrm{d}\tilde{\pi}}(\mathbf{u}\mid\mathbf{x}).
\end{equation}
Under this reparametrization, both the expected cost and the KL divergence term can be expressed as integrals with respect to $\tilde{\pi}$. Thus, the Bellman objective as in Eq.~\eqref{eq:bellman_app} reduces to
\begin{equation}
\int_{\mathcal{U}}\mathcal{Q}(\mathbf{x},\mathbf{u})\,p(\mathbf{u})\,\mathrm{d}\tilde{\pi}(\mathbf{u})
+\lambda\int_{\mathcal{U}}p(\mathbf{u})\log p(\mathbf{u})\,\mathrm{d}\tilde{\pi}(\mathbf{u}),
\label{eq:lag_obj_app}
\end{equation}
subject to the normalization constraint $\int_{\mathcal{U}}p(\mathbf{u})\,\mathrm{d}\tilde{\pi}(\mathbf{u})=1$.
To enforce this constraint, we introduce a Lagrange multiplier $\alpha$ and formulate the associated Lagrangian functional
\begin{align}
\mathcal{L}(p)=\int_{\mathcal{U}}\!\bigl[\mathcal{Q}(\mathbf{x},\mathbf{u})p(\mathbf{u})+\lambda p(\mathbf{u})\log p(\mathbf{u})\bigr]\,\mathrm{d}\tilde{\pi}
\notag \\
-\alpha\!\left(\int_{\mathcal{U}}p(\mathbf{u})\,\mathrm{d}\tilde{\pi}-1\right).
\end{align}

We now consider a perturbed density of the form $p_{\varepsilon}(\mathbf{u})=p(\mathbf{u})+\varepsilon h(\mathbf{u})$,
where $h$ is an arbitrary variation satisfying $\int_{\mathcal{U}} h\,\mathrm{d}\tilde{\pi}=0$.
Computing the first-order variation of $\mathcal{L}$ along this perturbation yields
\begin{equation}
\frac{\mathrm{d}}{\mathrm{d}\varepsilon}\mathcal{L}(p_\varepsilon)\Big|_{\varepsilon=0}
=\int_{\mathcal{U}}\!\bigl[\mathcal{Q}(\mathbf{x},\mathbf{u})+\lambda\bigl(1+\log p(\mathbf{u})\bigr)-\alpha\bigr]\,h(\mathbf{u})\,\mathrm{d}\tilde{\pi}.
\end{equation}
A necessary condition for optimality is that the first variation vanishes for every admissible perturbation $h$. By the fundamental lemma of the calculus of variations, this requires the integrand to vanish $\tilde{\pi}$-almost everywhere on $\mathcal{U}$, i.e,
\begin{equation}
\mathcal{Q}(\mathbf{x},\mathbf{u})
+
\lambda\bigl(1+\log p(\mathbf{u})\bigr)
-\alpha
=0.
\end{equation}
Solving this equation for $p(\mathbf{u})$ produces the unnormalized Boltzmann form
\begin{equation}
p(\mathbf{u})=\exp\!\Bigl(\frac{\alpha}{\lambda}-1\Bigr)\exp\!\Bigl(-\frac{\mathcal{Q}(\mathbf{x},\mathbf{u})}{\lambda}\Bigr).
\end{equation}

The remaining constant is determined by enforcing the normalization condition $\int_{\mathcal{U}} p\,\mathrm{d}\tilde{\pi}=1$, which yields the unique minimizer
\begin{equation}
\pi^*(\mathbf{u}|\mathbf{x})
=\frac{e^{-\mathcal{Q}(\mathbf{x},\mathbf{u})/\lambda}\,\tilde{\pi}(\mathbf{u}|\mathbf{x})}
{\displaystyle\int_{\mathcal{U}}e^{-\mathcal{Q}(\mathbf{x},\mathbf{v})/\lambda}\,\tilde{\pi}(\mathbf{v}|\mathbf{x})\,\mathrm{d}\mathbf{v}}.
\label{eq:pistar_app}
\end{equation}

\textbf{Step 2: Establishing the variational equality.}

We proceed to establish the variational equality by relating the objective evaluated at an arbitrary policy $\pi$ to that at the optimal policy $\pi^*$. Define the regularized cost functional
\begin{equation}
\begin{aligned}
I^{\pi}
&:= \mathbb{E}_{\pi}[\mathcal{Q}(\mathbf{x},\mathbf{u})]
   + \lambda D_{\mathrm{KL}}(\pi\|\tilde{\pi}) \\
&= \int_{\mathcal{U}}\mathcal{Q}(\mathbf{x},\mathbf{u})\,\mathrm{d}\pi
 + \lambda \int_{\mathcal{U}}
 \log\!\Bigl(\frac{\mathrm{d}\pi}{\mathrm{d}\tilde{\pi}}\Bigr)\,\mathrm{d}\pi .
\end{aligned}
\label{eq:Ipi_app}
\end{equation}
The key step is to apply the chain rule for Radon--Nikodym derivatives,
$\frac{\mathrm{d}\pi}{\mathrm{d}\tilde{\pi}}=\frac{\mathrm{d}\pi}{\mathrm{d}\pi^{*}}\frac{\mathrm{d}\pi^{*}}{\mathrm{d}\tilde{\pi}}$,
which, upon taking logarithms, gives the decomposition
\begin{equation}
\log\!\Bigl(\frac{\mathrm{d}\pi}{\mathrm{d}\tilde{\pi}}\Bigr)
=\log\!\Bigl(\frac{\mathrm{d}\pi}{\mathrm{d}\pi^{*}}\Bigr)+\log\!\Bigl(\frac{\mathrm{d}\pi^{*}}{\mathrm{d}\tilde{\pi}}\Bigr).
\end{equation}

From the explicit form of $\pi^*$ given in Eq.~\eqref{eq:pistar_app}, we obtain
\begin{equation}
\log\!\Bigl(\frac{\mathrm{d}\pi^{*}}{\mathrm{d}\tilde{\pi}}(\mathbf{u}|\mathbf{x})\Bigr)
=-\frac{1}{\lambda}\mathcal{Q}(\mathbf{x},\mathbf{u})
-\log\!\int_{\mathcal{U}}e^{-\mathcal{Q}(\mathbf{x},\mathbf{v})/\lambda}\,\mathrm{d}\tilde{\pi}(\mathbf{v}|\mathbf{x}).
\end{equation}
Substituting this expression into Eq.~\eqref{eq:Ipi_app} and denoting the log-partition function by
$Z:=\int_{\mathcal{U}}e^{-\mathcal{Q}/\lambda}\,\mathrm{d}\tilde{\pi}$, we have
\begin{align}
I^{\pi}
&\!=\!\int_{\mathcal{U}}\mathcal{Q}\,\mathrm{d}\pi
\!+\!\lambda\int_{\mathcal{U}}\log\!\Bigl(\frac{\mathrm{d}\pi}{\mathrm{d}\pi^{*}}\Bigr)\,\mathrm{d}\pi
\!+\!\lambda\int_{\mathcal{U}}\Bigl[-\frac{1}{\lambda}\mathcal{Q}\!-\!\log Z\Bigr]\,\mathrm{d}\pi 
\notag\\
&=-\lambda\log\!\int_{\mathcal{U}}e^{-\mathcal{Q}(\mathbf{x},\mathbf{u})/\lambda}\,\mathrm{d}\tilde{\pi}
+\lambda D_{\mathrm{KL}}(\pi\|\pi^{*}).
\label{eq:Ipi_final_app}
\end{align}
Since \(D_{\mathrm{KL}}(\pi\|\pi^{*})\ge 0\), with equality if and only if
\(\pi=\pi^{*}\) \(\tilde{\pi}\)-almost surely, taking the infimum over all
admissible policies \(\pi\) in Eq.~\eqref{eq:Ipi_final_app} establishes the
variational equality in Eq.~\eqref{eq:variational} and confirms that
\(\pi^*\) is its unique minimizer.

\textbf{Step 3: Gaussian closure in the linear--quadratic setting.}

Finally, we verify that \(\pi^*\) admits a closed-form Gaussian representation in the linear--quadratic case. When \(\mathcal{Q}(\mathbf{x},\mathbf{u})\) is quadratic in \(\mathbf{u}\), the term \(\exp(-\mathcal{Q}/\lambda)\) forms a Gaussian kernel in \(\mathbf{u}\). If the reference policy \(\tilde{\pi}(\mathbf{u}|\mathbf{x})\) is also Gaussian, then \(e^{-\mathcal{Q}/\lambda}\tilde{\pi}\) remains Gaussian. Consequently, \(\pi^*\) defined in Eq.~\eqref{eq:pistar_app} is Gaussian, with mean and covariance computable in closed form. This Gaussian closure property underpins the analytical characterization of the feedback Nash equilibrium in Theorem~\ref{th1}. This completes the proof.
\label{lemma1_proof}
\end{proof}
\subsection{Proof of Theorem~\ref{th1}}
\label{sec-proof-th1}

\begin{proof}
We proceed by backward induction on $t$, showing that $V_t^{i*}$ retains a
quadratic form at every time step and deriving the explicit recursion for its coefficients.

\textbf{Base case ($t=T$).}
At the terminal time, the value function is prescribed by the terminal cost,
$V_T^i(\mathbf{x}_T)=\phi^i(\mathbf{x}_T)$, where
\begin{equation}
    \phi^i(\mathbf{x}_T)=\frac{1}{2}(\mathbf{x}_T-\mathbf{x}_T^{\mathrm{ref}})^\top Q_T^i(\mathbf{x}_T-\mathbf{x}_T^{\mathrm{ref}}),
\qquad Q_T^i \succeq 0.
\end{equation}
This is quadratic, so the claimed structure holds at \(t=T\). In deviation coordinates
\(\delta\mathbf{x}_T=\mathbf{x}_T-\bar{\mathbf{x}}_T\), the boundary conditions are
\(Z_T^i=Q_T^i\) and \(z_T^i=0\), provided \(\bar{\mathbf{x}}_T=\mathbf{x}_T^{\mathrm{ref}}\).

\textbf{Induction step.}
Suppose the quadratic structure persists at time $t+1$, namely
\begin{equation}
V_{t+1}^{i}(\mathbf{x}_{t+1})
=\tfrac{1}{2}\mathbf{x}_{t+1}^\top Z_{t+1}^{i}\mathbf{x}_{t+1}
+z_{t+1}^{i\top}\mathbf{x}_{t+1}+\eta_{t+1}^{i}.
\label{eq:Vt1_quad}
\end{equation}

Since \(c_t^i\) is quadratic by Eq.~\eqref{eq:5} and the linearized dynamics in Eq.~\eqref{eq:4} are affine, composing Eq.~\eqref{eq:Vt1_quad} with the dynamics shows that \(\mathcal{Q}_t^i = c_t^i + V_{t+1}^i(\mathbf{x}_{t+1})\) is quadratic in \((\delta\mathbf{x}_t,\mathbf{u}_t)\). By Lemma~\ref{lemma:variational}, the Bellman update is
\begin{equation}
V_t^i(\mathbf{x}_t)
=-\lambda^i\log\!\int\exp\!\Bigl(-\tfrac{1}{\lambda^i}\mathcal{Q}_t^i(\mathbf{x}_t,\mathbf{u}_t)\Bigr)\,\mathrm{d}\tilde{\pi}_t^i.
\end{equation}
The Gaussian closure property (Step~3 of Lemma~\ref{lemma:variational}) ensures that the log-partition function is quadratic in its natural parameters, which depend affinely on $\mathbf{x}_t$. Hence $V_t^{i*}$ admits the quadratic representation
\begin{equation}
V_t^i(\mathbf{x}_t)
=\tfrac{1}{2}\mathbf{x}_t^\top Z_t^i\mathbf{x}_t+z_t^{i\top}\mathbf{x}_t+\bar{\eta_t}^i.
\label{eq:value_quad_app}
\end{equation}

It remains to obtain the explicit coefficients \((Z_t^i,z_t^i)\) by expanding the Bellman equation, optimizing over \((\mu_t^i,\Sigma_t^i)\), and matching the coefficients of \(\delta\mathbf{x}_t\). Substituting Eq.~\eqref{eq:bellman_app} and the quadratic cost approximation yields
\begin{align}
&V_t^{i*}(\mathbf{x}_t)
=
\min_{\pi_t^i}\mathbb{E}^{\pi}\Bigg[\underbrace{
c_t^i(\bar{\mathbf{x}}_t,\bar{\mathbf{u}}_t,\bm{\omega})
+\tfrac{1}{2}(\delta\mathbf{x}_t^\top Q_t^i+2\bm{\ell}_t^{i\top})\delta\mathbf{x}_t}_{\text{(A)}}
\notag\\
&\quad+
\underbrace{
\tfrac{1}{2}\sum_{j\in[N]}
(\delta\mathbf{u}_t^{j\top}R_t^{ij}+2\bm{r}_t^{ij\top})\delta\mathbf{u}_t^j
}_{\text{(A) Stage cost}}
+
\underbrace{
\lambda^i D_{\mathrm{KL}}\!\left(\pi_t^i\,\|\,\tilde{\pi}_t^i\right)
}_{\text{(B) KL divergence}}
\notag\\
&\quad+
\underbrace{
\mathbb{E}^{\mathbf{x}_{t+1}}\!\left[
V_{t+1}^{i*}\!\left(
A_t\delta\mathbf{x}_t+\sum_{j\in[N]}B_t^j\delta\mathbf{u}_t^j
\right)
\right]
}_{\text{(C) Next-step value}}
\Bigg].
\label{eq:bellman_expanded}
\end{align}
Since each agent uses a Gaussian policy \(\pi_t^i\sim\mathcal{N}(\mu_t^i,\Sigma_t^i)\), we write
\(\delta\mathbf{u}_t^i=\mu_t^i+\epsilon_t^i\), where
\(\epsilon_t^i\sim\mathcal{N}(0,\Sigma_t^i)\). We then evaluate the expectation of each bracketed term.

\paragraph{Evaluating Term (1)}
Taking expectations with respect to the Gaussian policy yields
\begin{align}
\text{(A)}
=\,&c_t^i(\bar{\mathbf{x}}_t,\bar{\mathbf{u}}_t,\bm{\omega})
+\tfrac{1}{2}(\delta\mathbf{x}_t^\top Q_t^i+2\bm{\ell}_t^{i\top})\delta\mathbf{x}_t \notag\\
&+\tfrac{1}{2}\sum_{j\in[N]}\!\Bigl(\mu_t^{j\top}R_t^{ij}\mu_t^j
+\operatorname{tr}(R_t^{ij}\Sigma_t^j)+2\bm{r}_t^{ij\top}\mu_t^j\Bigr).
\label{eq:term1_app}
\end{align}

\paragraph{Evaluating Term (2)}
The closed-form KL divergence between the two Gaussians gives
\begin{align}
\text{(B)}
=\,&\frac{\lambda^i}{2}\Bigl(n_u^i
-\log\det(\Sigma_t^i)+\log\det(\tilde{\Sigma}_t^i)
+\operatorname{tr}\bigl((\tilde{\Sigma}_t^i)^{-1}\Sigma_t^i\bigr)
\notag \\
&+(\mu_t^i-\tilde{\mu}_t^i)^\top(\tilde{\Sigma}_t^i)^{-1}(\mu_t^i-\tilde{\mu}_t^i)\Bigr).
\label{eq:term2_app}
\end{align}

\paragraph{Evaluating Term (C)}
Substituting Eq.~\eqref{eq:Vt1_quad} into the next-step value, computing the Gaussian moments, and separating purely state-dependent contributions into the quadratic and linear parts of $V_t^{i*}$, we obtain
\begin{align}
\text{(C)}
&=\tfrac{1}{2}\Bigl(\textstyle\sum_{j}B_t^j\mu_t^j\Bigr)^\top
Z_{t+1}^i\Bigl(\textstyle\sum_{j}B_t^j\mu_t^j\Bigr) \notag\\
&\quad+\sum_{j}(A_t\delta\mathbf{x}_t)^\top Z_{t+1}^iB_t^j\mu_t^j+\tfrac{1}{2}\operatorname{tr}\!\Bigl(Z_{t+1}^i\textstyle\sum_{j}B_t^{j\top}\Sigma_t^jB_t^j\Bigr) \notag\\
&\quad+\sum_{j}z_{t+1}^{i\top}B_t^j\mu_t^j + \mathrm{const}.
\label{eq:term3_app}
\end{align}

Aggregating Eqs.~\eqref{eq:term1_app}--\eqref{eq:term3_app} and discarding all terms independent of
$(\mu_t^i,\Sigma_t^i)$, the optimization problem reduces to minimizing the following functional
\begin{align}
\tilde{V}&_t^i
=\tfrac{1}{2}\mu_t^{i\top}R_t^{ii}\mu_t^i
\!+\!r_t^{ii\top}\mu_t^i
+\tfrac{\lambda^i}{2}(\mu_t^i-\tilde{\mu}_t^i)^\top(\tilde{\Sigma}_t^i)^{-1}(\mu_t^i-\tilde{\mu}_t^i)
\notag\\
&\quad+\tfrac{1}{2}\Bigl(A_t\delta\mathbf{x}_t+\textstyle\sum_{j}B_t^j\mu_t^j\Bigr)^\top
Z_{t+1}^i\Bigl(A_t\delta\mathbf{x}_t+\textstyle\sum_{j}B_t^j\mu_t^j\Bigr)
\notag\\
&\quad
+\operatorname{tr}\bigl((\tilde{\Sigma}_t^i)^{-1}\Sigma_t^i\bigr)\Bigr)
\!+\!\tfrac{1}{2}\operatorname{tr}(R_t^{ii}\Sigma_t^i)
\!+\!\tfrac{1}{2}\operatorname{tr}\bigl(Z_{t+1}^iB_t^i\Sigma_t^iB_t^{i\top}\bigr)
\notag\\
&\quad
+z_{t+1}^{i\top}\textstyle\sum_{j}B_t^j\mu_t^j
\tfrac{\lambda^i}{2}\Bigl(-\log\det(\Sigma_t^i)+\log\det(\tilde{\Sigma}_t^i).
\label{eq:Vtilde_app}
\end{align}

We now determine the optimal policy parameters $\mu_t^{i*}$ and $\Sigma_t^{i*}$ via the first-order necessary conditions.
\paragraph{Step 1. First-order condition with respect to $\mu_t^i$}
Differentiating Eq.~\eqref{eq:Vtilde_app} with respect to $\mu_t^i$ yields
\begin{align}
\partial \tilde{V}_t^i / &\partial{\mu_t^i}
=R_t^{ii}\mu_t^i+r_t^{ii}
+\lambda^i(\tilde{\Sigma}_t^i)^{-1}(\mu_t^i-\tilde{\mu}_t^i)
\notag\\
&\quad
+B_t^{i\top}Z_{t+1}^i\!\Bigl(A_t\delta\mathbf{x}_t+\textstyle\sum_{j}B_t^j\mu_t^j\Bigr)
+B_t^{i\top}z_{t+1}^i.
\label{eq:dV_dmu_app}
\end{align}
Setting Eq.~\eqref{eq:dV_dmu_app} equal to zero furnishes the stationarity condition. Since the resulting equation is linear in $(\mu_t^i,\delta\mathbf{x}_t)$, the optimal control mean is necessarily
affine in the state deviation, motivating the linear feedback ansatz
\begin{equation}
\mu_t^{i*}=-P_t^i\delta\mathbf{x}_t-\alpha_t^i.
\end{equation}
Substituting $\mu_t^{j*}=-P_t^j\delta\mathbf{x}_t-\alpha_t^j$ for every $j\in[N]$ into the stationarity condition and equating the coefficients of $\delta\mathbf{x}_t$
and the constant terms separately recovers the gain Eq.~\eqref{13a} and the offset Eq.~\eqref{13b}, respectively.
\paragraph{Step 2. First-order condition with respect to $\Sigma_t^i$}
Differentiating Eq.~\eqref{eq:Vtilde_app} with respect to the covariance matrix $\Sigma_t^i$ gives
\begin{align}
\frac{\partial \tilde{V}_t^i}{\partial \Sigma_t^i}
&= \!
\frac{\lambda^i}{2}\bigl(-(\Sigma_t^i)^{-1}\!\!+\!(\tilde{\Sigma}_t^i)^{-1}\bigr)
\!\!+\!\!\tfrac{1}{2}R_t^{ii}\!
+\!\tfrac{1}{2}B_t^{i\top}Z_{t+1}^iB_t^i .
\label{eq:dV_dSigma_app}
\end{align}
Setting the first-order condition 
$\partial \tilde{V}_t^i / \partial \Sigma_t^i = 0$
and solving for $\Sigma_t^i$ yield the closed-form optimal covariance
\begin{equation}
\Sigma_t^{i*}
=
\Bigl[
\tfrac{1}{\lambda^i}
\bigl(R_t^{ii}+B_t^{i\top}Z_{t+1}^iB_t^i\bigr)
+
(\tilde{\Sigma}_t^i)^{-1}
\Bigr]^{-1}.
\end{equation}
\paragraph{Step 3. Backward Riccati recursion for $(Z_t^i,z_t^i)$}
Substituting the optimal policies $\mu_t^{j*}=-P_t^j\delta\mathbf{x}_t-\alpha_t^j$
and $\Sigma_t^{i*}$ back into Eq.~\eqref{eq:bellman_expanded} and collecting
quadratic and linear terms in $\delta\mathbf{x}_t$, we introduce the closed-loop quantities
\begin{equation}
F_t=A_t-\sum_{j\in[N]}B_t^jP_t^j,\qquad\beta_t=-\sum_{j\in[N]}B_t^j\alpha_t^j,
\end{equation}
with all remaining scalar constants absorbed into $\bar{\eta_t}^i$.
Identifying the resulting expression with the postulated quadratic form in Eq.~\eqref{eq:value_quad_app} and matching the quadratic and linear coefficients of \(\delta\mathbf{x}_t\) directly yields the recursions for \(Z_t^i\) and \(z_t^i\), which coincide with Eqs.~\eqref{14a} and~\eqref{14b} in the main text. This completes the proof.
\end{proof}
\subsection{Proof of Proposition~\ref{prop:lambda_wellposed}}
\label{proof_pro2}
\begin{proof}

We first prove Statement~1) by backward induction on $t$. The terminal
condition of the value recursion satisfies
\begin{equation}
    Z_T^i = Q_T^i \succeq 0,\qquad \forall i\in[N],
\end{equation}
by the positive semidefiniteness of the terminal quadratic approximation.

Suppose that, for some $t\in\{0,\ldots,T-1\}$, the induction hypothesis
holds:
\begin{equation}
    Z_{t+1}^i\succeq0,\qquad \forall i\in[N].
\end{equation}
By Assumption~\ref{ass:regularity}~(A4), the coupled linear systems in
Eqs.~\eqref{13a}--\eqref{13b} are nonsingular. Hence, the feedback gains
$\{P_t^i,\alpha_t^i\}_{i\in[N]}$ are uniquely determined. Substituting these gains into Eq.~\eqref{14a}, each term in the recursion
is positive semidefinite:
$Q_t^i\succeq0$ by the local quadratic approximation,
$P_t^{j\top}R_t^{ij}P_t^j\succeq0$ since $R_t^{ij}\succeq0$, and
$F_t^\top Z_{t+1}^iF_t\succeq0$ by the induction hypothesis. Moreover,
Eqs.~\eqref{eq:lambda_state} and~\eqref{eq:frozen_lambda} imply
$\lambda_t^{i,k}\ge\lambda_{\min}^i>0$, while
$\tilde{\Sigma}_t^i\succ0$ follows from the positive-definiteness
condition stated in Proposition~\ref{prop:lambda_wellposed}. Hence, we have
$
\lambda_t^{i,k}P_t^{i\top}(\tilde{\Sigma}_t^i)^{-1}P_t^i\succeq0.
$
Therefore, we get $Z_t^i\succeq0$. This completes the backward induction and proves Statement~1).

We next prove Statement~2). For any $t=0,\ldots,T-1$, Statement~1)
implies $Z_{t+1}^i\succeq0$. Therefore,
\begin{equation}
    B_t^{i\top}Z_{t+1}^iB_t^i\succeq0.
\end{equation}
Using again $\lambda_t^{i,k}\ge\lambda_{\min}^i>0$ and
$(\tilde{\Sigma}_t^i)^{-1}\succ0$, it follows that
\begin{equation}
    \lambda_t^{i,k}(\tilde{\Sigma}_t^i)^{-1}
    \succeq
    \lambda_{\min}^i
    \sigma_{\min}\!\left((\tilde{\Sigma}_t^i)^{-1}\right) I .
\end{equation}
Together with the strong convexity condition
$R_t^{ii}\succeq r_{\min}^iI\succ0$, it follows that
\begin{align}
    M_t^{i,k}
    &=
    R_t^{ii}
    +\lambda_t^{i,k}(\tilde{\Sigma}_t^i)^{-1}
    +B_t^{i\top}Z_{t+1}^iB_t^i \notag\\
    &\succeq
    \left(
    r_{\min}^i+
    \lambda_{\min}^i
    \sigma_{\min}\!\left((\tilde{\Sigma}_t^i)^{-1}\right)
    \right)I 
    \succeq
    m_0^iI\succ0 .
\end{align}
Therefore, $M_t^{i,k}$ is uniformly positive definite in $t$ and $k$,
which proves Statement~2).

For Statement~3), substituting the frozen coefficient $\lambda_t^{i,k}$
into Eq.~\eqref{conv1} and using the definition of $M_t^{i,k}$ gives
\begin{equation}
    \Sigma_t^{i*}
    =
    \lambda_t^{i,k}\left(M_t^{i,k}\right)^{-1}.
\end{equation}
Since $\lambda_t^{i,k}>0$ and $M_t^{i,k}\succ0$, it follows that
$\Sigma_t^{i*}\succ0$. Thus, the covariance update is well-defined.

Finally, since the coupled linear system in Eqs.~\eqref{13a}--\eqref{13b}
is nonsingular, the feedback gains $\{P_t^i,\alpha_t^i\}_{i\in[N]}$ are
unique, and so is the mean
$\mu_t^{i*}=-P_t^i\delta\mathbf{x}_t-\alpha_t^i$. Together with the
uniquely defined positive definite covariance $\Sigma_t^{i*}$, this gives
a unique Gaussian policy and thus a unique Gaussian FBNE for the frozen
local game. This completes the proof.
\end{proof}
\subsection{Proof of Proposition~\ref{prop:lambda_regular}}
\label{proof_pro1}
\begin{proof}
We prove the Lipschitz continuity of $\lambda^i$ by composing the
Lipschitz continuity of $d_{\mathrm{obs}}^i$, the smooth Gaussian kernel,
and the affine transformation defining $\lambda^i$.

\textit{Step 1. ($d_{\mathrm{obs}}^i$ is $L_d$-Lipschitz).}
Let $p^i=\mathcal{P}^i\mathbf{x}$ and
$p'^i=\mathcal{P}^i\mathbf{x}'$. By Assumption~\ref{ass:regularity}~(A3),
the position extraction map $\mathcal{P}^i$ is linear and non-expansive. Hence,
\begin{equation}
\|p^i-p'^i\|
=
\|\mathcal{P}^i(\mathbf{x}-\mathbf{x}')\|
\le
\|\mathbf{x}-\mathbf{x}'\|.
\end{equation}
For the static obstacle term, the distance function to a non-empty closed
set is $1$-Lipschitz~\cite{rudin1976principles}. Therefore,
\begin{equation}
\big|
\mathrm{dist}(p^i,\mathcal{O})
-
\mathrm{dist}(p'^i,\mathcal{O})
\big|
\le
\|p^i-p'^i\|
\le
\|\mathbf{x}-\mathbf{x}'\|.
\end{equation}
For each interaction-distance term with $j\neq i$, the reverse triangle
inequality gives
\begin{align}
\big|
\|p^i-p^j\|
-
\|p'^i-p'^j\|
\big|
&\le
\|(p^i-p^j)-(p'^i-p'^j)\| \notag\\
&\le
\|p^i-p'^i\|+\|p^j-p'^j\| \notag\\
&\le
2\|\mathbf{x}-\mathbf{x}'\|.
\end{align}
Since the pointwise minimum of finitely many Lipschitz functions remains
Lipschitz, $d_{\mathrm{obs}}^i$ is Lipschitz continuous. In particular,
we may take
\begin{equation}
\big|
d_{\mathrm{obs}}^i(\mathbf{x})
-
d_{\mathrm{obs}}^i(\mathbf{x}')
\big|
\le
L_d\|\mathbf{x}-\mathbf{x}'\|,
\qquad
L_d=2 .
\end{equation}

\textit{Step 2. ($\rho^i$ is $L_\rho$-Lipschitz).}
Since the function $r\mapsto \exp(-r^2/(2\sigma^2))$ is continuously
differentiable with bounded derivative on $[0,\infty)$, it is Lipschitz
continuous. Together with the $L_d$-Lipschitz continuity of
$d_{\mathrm{obs}}^i$, this implies
\begin{equation}
|\rho^i(\mathbf{x})-\rho^i(\mathbf{x}')|
\le
L_\rho\|\mathbf{x}-\mathbf{x}'\|,
\end{equation}
for some constant $L_\rho>0$.

\textit{Step 3. ($\lambda^i$ is $C_\lambda^i$-Lipschitz).}
Since $\lambda^i$ is an affine function of $\rho^i$ in
Eq.~\eqref{eq:lambda_state}, the Lipschitz continuity of $\rho^i$ gives
\begin{equation}
|\lambda^i(\mathbf{x})-\lambda^i(\mathbf{x}')|
\le
C_\lambda^i\|\mathbf{x}-\mathbf{x}'\|,
C_\lambda^i=(\lambda_{\max}^i-\lambda_{\min}^i)L_\rho .
\end{equation}
This completes the proof.
\end{proof}

\bibliographystyle{IEEEtran}
\bibliography{references}
\end{document}